\documentclass[aos,preprint]{imsart_arxiv}
\usepackage{amsmath}
\usepackage{amsfonts}
\usepackage{amssymb}
\usepackage{amsfonts}
\usepackage{amssymb}
\usepackage{amsthm}
\usepackage[pdftex]{graphicx}
\usepackage{enumerate}
\usepackage[numbers]{natbib}
\usepackage[colorlinks,citecolor=blue,urlcolor=blue]{hyperref}
\usepackage{caption}
\usepackage{subcaption}
\RequirePackage[OT1]{fontenc}
\allowdisplaybreaks

\startlocaldefs
\newcommand{\argmin}{\operatornamewithlimits{arg \,min}}
\newcommand{\argmax}{\operatornamewithlimits{arg \,max}}
\numberwithin{equation}{section}
\theoremstyle{plain}
\newtheorem{thm}{Theorem}
\newtheorem{lem}{Lemma}
\newtheorem{prop}{Proposition}
\newtheorem{cor}{Corollary}
\begin{document}

\begin{frontmatter}
\title{Orthogonal symmetric non-negative matrix factorization under the stochastic block model}
\runtitle{Orthogonal symmetric non-negative matrix factorization}
\thankstext{T1}{Supported in part by NSF Grant DMS-1406455.}




\author{\fnms{Subhadeep} \snm{Paul}\ead[label=e1]{spaul10@illinois.edu}}
\and
\author{\fnms{Yuguo} \snm{Chen}\ead[label=e2]{yuguo@illinois.edu}}
\affiliation{University of Illinois at Urbana-Champaign}

\runauthor{S. Paul and Y. Chen}

\address{Department of Statistics\\ University of Illinois at Urbana-Champaign\\
Champaign, IL 61820\\ USA \\
\printead{e1}\\
\phantom{E-mail:\ }\printead*{e2}
}

\begin{abstract}
We present a method based on the orthogonal symmetric non-negative matrix tri-factorization of the normalized Laplacian matrix for community detection in complex networks. While the exact factorization of a given order may not exist and is NP hard to compute, we obtain an approximate factorization by solving an optimization problem. We establish the connection of the factors obtained through the factorization to a non-negative basis of an invariant subspace of the estimated matrix, drawing parallel with the spectral clustering. Using such factorization for clustering in networks is motivated by analyzing a block-diagonal Laplacian matrix with the blocks representing the connected components of a graph. The method is shown to be consistent for community detection in graphs generated from the stochastic block model and the degree corrected stochastic block model. Simulation results and real data analysis show the effectiveness of these methods under a wide variety of situations, including sparse and highly heterogeneous graphs where the usual spectral clustering is known to fail. Our method also performs better than the state of the art in popular benchmark network datasets, e.g., the political web blogs and the karate club data.
\end{abstract}

\begin{keyword}[class=MSC]
\kwd[Primary ]{62F12, 62H30}
\kwd[; secondary ]{90B15, 15A23.}
\end{keyword}

\begin{keyword}
\kwd{Community detection}
\kwd{invariant subspaces}
\kwd{network data}
\kwd{non-negative matrix factorization}
\kwd{stochastic block model}
\end{keyword}

\end{frontmatter}

\section{Introduction}
Over the last two decades there has been an enormous increase in literature on statistical inference of network data motivated by their ever increasing applications in computer science, biology and economics. A network consists of a set of entities called nodes or vertices and a set of connections among them called edges or relations. While there are many interesting statistical problems associated with network data, one problem that has received considerable attention in literature, particularly over the last decade, is the problem of detecting communities or clusters of nodes in a network. A community is often defined as a group of nodes which are more ``similar" to each other as compared to the rest of the network. A common notion of such similarity is structural similarity, whereby a community of nodes are more densely connected among themselves than they are to the rest of the network.

Several methods have been proposed in the literature for efficient detection of network communities. Maximizing a quality function for community structure called ``modularity" has been shown to perform quite well in a wide variety of networks \citep{ng04}. Both the Newman-Girvan modularity and the likelihood modularity (i.e., the modularity based on the model's likelihood function) were shown to be consistent under the stochastic block model (SBM) in \citet{bc09} and under the degree corrected stochastic block model (DCSBM) in \citet{zlz12}. Community detection using spectral clustering and its variants have also been studied extensively under both SBM and DCSBM \citep{rcy11,qr13,j15,lei14}.

In this paper we consider methods for community detection in networks based on the non-negative matrix factorization of the Laplacian matrix of the network. Non-negative matrix factorization (NMF) has received considerable attention in the machine learning and data mining literature since it was first introduced in \citet{lee99}. The method has many good properties in terms of performance and interpretability, and is extremely popular in many applications including image and signal processing, information retrieval, document clustering, neuroscience and bio-informatics. A matrix $X$ is said to be non-negative if all its elements are non-negative, i.e., $X_{ij} \geq 0$ for all $i,j$. The general NMF of order $K$ decomposes a non-negative matrix $X \in \mathcal{R}^{N\times M}_{+}$ into two non-negative factor matrices $W \in \mathcal{R}^{N\times K}_{+}$ and $H \in \mathcal{R}^{K\times M}_{+}$, i.e., $X=WH$. When $K \leq \min \{M,N\}$, NMF can also be looked upon as a dimension reduction technique that ``decomposes a matrix into parts" that generate it \citep{lee99}.

However, an exact NMF of order $K$ may not exist for any given non-negative matrix and even if one does, finding the exact NMF in general settings is a computationally difficult problem and has been shown to be NP hard previously \citep{vavasis09}. In fact it was shown that, not just finding an exact order $K$ NMF, but also verifying the existence of the same is NP hard. To remedy this, several algorithms for an approximate solution has been proposed in the literature \citep{lee01,lin07,cich09}. Popular optimization based algorithms aims to minimize the difference between $X$ and $WH$ in Frobenius norm under the non-negativity constraints.  However, a natural question arises that given the matrix $X$ is generated by exact multiplication of non-negative matrices (the ``parts"), whether the decomposition can uniquely identify those parts of the generative model. A number of researchers have tackled this problem both geometrically and empirically \citep{donoho03,hoyer04,laurberg08,huang14}.

NMF has also been applied in the context of clustering \citep{xu03,ding05,ding06,kim08}. The ``low-rank" NMF, where $K \leq \min \{M,N\}$, can be used to obtain a low-dimensional factor matrix, which can subsequently be used for clustering.  \citet{ding05} showed interesting connections of NMF with other clustering algorithms such as kernel k-means and spectral clustering. For applications in graph clustering where we generally have a symmetric adjacency matrix or a Laplacian matrix as the non-negative matrix, a symmetric version of the factorization was proposed in \citet{wang11}. This factorization, called the symmetric non-negative matrix factorization (SNMF), has been empirically shown to yield good results in various clustering scenarios, including community detection in networks \citep{wang11,kuang12}. \citet{arora11} used a special case of SNMF, the left stochastic matrix factorization, for clustering, and derived perturbation bounds. \citet{yang12} used a regularized version of the SNMF algorithm for clustering, while \citet{psorakis11} used a Bayesian NMF for overlapping community detection. In this paper we consider another non-negative matrix factorization designed to factorize symmetric matrices, the orthogonal symmetric non-negative matrix (tri) factorization (OSNTF) \citep{ding06,pompili14}.

We use the normalized graph Laplacian matrix as our non-negative matrix for factorization instead of the usual adjacency matrix, as it has been recently shown to provide better clustering quality in spectral clustering for graphs generated from the SBM \citep{sarkar15}. In contrast with earlier approaches, the requirement of being orthogonal in OSNTF adds another layer of extra constraints, but generates sparse factors which are good for clustering. It also performs well in our experiments. We prove that OSNTF is consistent under both the stochastic block model and its degree corrected variant. Through simulations and real data examples we demonstrate the efficacy of both OSNTF and SNMF in community detection. In particular we show the advantages of proposed methods over the usual Laplacian based spectral clustering and its modifications in terms of regularization and projection within unit circle \citep{lei14,qr13}. The proposed methods do not require such modifications even when there is high degree heterogeneity in a sparse graph. An application to the widely analyzed political blogs data \citep{adamic05} results in a performance superior to the state of the art methods like SCORE \citep{j15}. The main contribution of the paper is in deriving theoretical results on consistency of community detection under SBM and DCSBM using a NMF based method.

The rest of the paper is organized as follows. Section 2 describes the two methods and the corresponding algorithms. Section 3 motivates the use of these methods for community detection. Section 4 describes consistency results for OSNTF under both SBM and DCSBM. Section 5 gives computational details and simulations. Section 6 reports application of the methods in real world network datasets and Section 7 gives concluding remarks.

\section{Methods and algorithms}

We consider an undirected graph $G$ on a set of $N$ vertices. The adjacency matrix $A$ associated with the graph is defined as a binary symmetric matrix with $A_{ij}=1$, if node $i$ and $j$ are connected and $A_{ij}=0$, if they are not. Throughout this paper we do not allow the graphs to have self loops and multiple edges. In this context we define degree of a node as the number of nodes it is connected to, i.e., $d_{i}=\sum_{j} A_{ij}$. The corresponding normalized graph Laplacian matrix can be defined as
\[
L=D^{-1/2}AD^{-1/2},
\]
where $D$ is a diagonal matrix with the degrees of the nodes as elements, i.e., $D_{ii}=d_{i}$. Throughout the paper $H \geq 0$ for a matrix means the matrix $H$ is non-negative, i.e., all its elements are non-negative. We denote the Frobenius norm as $\|\cdot\|_F$ and the spectral norm as $\|\cdot\|_2 $. We use $\|\cdot\|$ to denote the $L^2$ norm (Euclidean norm) of a vector.

We first describe SNMF which was previously used for community detection in networks using adjacency matrix by \citet{wang11} and \citet{kuang12}. Given a symmetric positive semi-definite adjacency matrix $A$ of a graph, the \textit{exact} SNMF of order $K$ for the adjacency matrix can be written as
\begin{equation}
A_{N\times N}=HH^{T}, \qquad H_{N \times K}\geq 0. \label{SNMF}
\end{equation}
However since finding or even verifying the existence of the exact SNMF is NP-hard, an approximate solution is obtained instead by solving the following optimization problem, which seeks to minimize the distance in Frobenius norm between $A$ and $HH^{T}$, i.e., we find,
\begin{equation}
\hat{H}= \argmin_{H_{N \times K}\geq 0} \|A-HH^{T}\|_{F}.
\label{SNMFoptim}
\end{equation}
Denoting $\hat{A}=\hat{H}\hat{H}^{T}$, it is easy to see that $\hat{H}$ is an exact SNMF factor of $\hat{A}$. We will refer to the solution of this optimization problem as SNMF. Clearly if $A$ has an exact factorization as in Equation (\ref{SNMF}), that factorization will be the solution to this optimization problem and then SNMF will refer to that exact factorization. The \textit{exact} SNMF of order $K$ for the normalized graph Laplacian matrix can be similarly defined as
\begin{equation}
L_{N \times N}=HH^{T}, \qquad H_{N \times K}\geq 0. \label{SNMFL}
\end{equation}

However since $HH^{T}$ is necessarily positive semi-definite, the \textit{exact} factorizations in Equations (\ref{SNMF}) and (\ref{SNMFL}) can not exist for matrices $A$ or $L$ that are not positive semi-definite. Moreover, being positive semi-definite is not a sufficient condition for the non-negative matrix $A$ to have a decomposition of the form $HH^{T}$ with $H\geq 0$. A non-negative positive semi-definite matrix is called doubly non-negative matrix. A doubly non-negative matrix that can be factorized into a SNMF is called a completely positive matrix \citep{berman03,gray80}. Despite these restrictions, we can still use this decomposition in practice by Equation (\ref{SNMFoptim}) since the optimization algorithms only tries to approximate the matrix $A$. However, obtaining theoretical guarantees on performance will be quite difficult.

In an attempt to remedy this situation, we consider another symmetric non-negative matrix factorization where the matrix $A$ is not required to be completely positive. Given an adjacency matrix $A$ of a graph this factorization, called the orthogonal symmetric non-negative matrix tri-factorization (OSNTF) of order $K$ \citep{ding06}, can be written as
\begin{equation}
A_{N \times N}=HSH^{T}, \qquad H_{N \times K}\geq 0, \, S_{K \times K}\geq 0, \, H^{T}H=I. \label{OSNTF}
\end{equation}
The matrix $S$ is symmetric but not necessarily diagonal and can have both positive and negative eigenvalues. Note that having the $S$ matrix gives the added flexibility of factorizing matrices which are not positive semi-definite and hence has negative eigenvalues.
In this connection it is worth mentioning that another symmetric tri-factorization was defined in \citet{ding05} without the orthogonality condition on the columns of $H$. However we keep this orthogonality condition as it leads to sparse factors and our experiments indicate that it leads to better performance both in simulations and in real networks. The OSNTF for normalized graph Laplacian is defined identically as in Equation (\ref{OSNTF}) with $A$ being replaced by $L$.

As before, in practice it is difficult to obtain or verify the existence of the \textit{exact} OSNTF in Equation (\ref{OSNTF}) for any given adjacency matrix. Hence to obtain an approximate decomposition, we minimize the distance in Frobenius norm between $A$ and $HSH^{T}$, i.e., we find
\begin{equation}
[\hat{H}, \hat{S}]= \argmin_{H_{N \times K}\geq 0, \,S_{K \times K}\geq 0, \, H^{T}H=I} \|A-HSH^{T}\|_{F}.
\label{OSNTFoptim}
\end{equation}
The solution to this optimization problem will be referred to as OSNTF of $A$. If an exact OSNTF of $A$ exists then this solution will coincide with the exact OSNTF.

There are several algorithms proposed in the literature to solve the optimization problems in Equations (\ref{SNMFoptim}) and (\ref{OSNTFoptim}). We use the algorithm due to \citet{wang11} for SNMF where Equation (\ref{SNMFoptim}) is solved through gradient descent and the update rules are given by
\begin{equation*}
H_{ik} \leftarrow H_{ik} \left( \frac{1}{2}+\frac{1}{2}\frac{(AH)_{ik}}{(HH^TH)_{ik}} \right),
\end{equation*}
for $i=1, \ldots ,N$ and $k=1, \ldots, K$. For OSNTF we use the update rules given in \citet{ding06} :
\begin{align*}
S_{ik} & \leftarrow S_{ik} \sqrt{\left( \frac{(H^TAH)_{ik}}{(H^THSH^TH)_{ik}} \right)}, \\
H_{ik} & \leftarrow H_{ik} \sqrt{\left( \frac{(AHS)_{ik}}{(HH^TAHS)_{ik}} \right)}.
\end{align*}

The matrix $H$ is used for community detection in both SNMF and OSNTF. After the algorithm converges, the community label for the $i$th node is obtained by assigning the $i$th row of $H$ to the column corresponding to its largest element, i.e., node $i$ is assigned to community $k$ if
\begin{equation}
k=\argmax_{j \in \{1,\ldots, K\}} H_{ij}.
\label{assignment}
\end{equation}
Here the rows of $H$ represent the nodes and the columns represent the communities. This way each node is assigned to one of the $K$ communities. The matrix $H$ can be thought of as a soft clustering for the nodes in the graph.

Since the optimization problems in both SNMF and OSNTF are non-convex and the algorithms described above have convergence guarantees only to a local minimum, a proper initialization is required. We use either the spectral clustering or a variant of it, the regularized spectral clustering \citep{qr13}, to initialize both algorithms. Our experiments indicate the final solution is not too sensitive to which of the two initializations is used, but the number of iterations needed to converge (and consequently time to converge) can vary depending upon which algorithm is used to initialize the methods. However our method does not require any regularization in terms of removing the high degree nodes or adding a small constant term to all nodes even for sparse graphs, as is often used with spectral clustering \citep{krzakala13}. We also do not require projecting the rows of $H$ into a unit circle, even when the degrees are heterogeneous, as is necessary for spectral clustering to perform well in such situations \citep{qr13}.

\subsection{Another characterization of approximate OSNTF}
We characterize the optimization problem of approximate OSNTF in (\ref{OSNTFoptim}) as a maximization problem which will help us later to bound the error of approximation.

Let us denote the subclass of $N \times K$ non-negative matrices whose columns are orthonormal as $\mathcal{H}_{+}^{N \times K}$. Given a feasible $H \in \mathcal{H}_{+}^{N \times K} $, the square of the objective function in the optimization problem in (\ref{OSNTFoptim}) can be written as
\begin{align*}
J &=  tr[(A-HSH^{T})^{T}(A-HSH^{T})] \\
&=  tr (AA-2SH^{T}AH + SS ).
\end{align*}
Now one can optimize for $S$ in the following way. Differentiating $J$ with respect to $S$, we get
\begin{equation*}
\frac{\partial J}{\partial S} \equiv -2H^{T}AH +2 S =0,
\end{equation*}
which implies $\hat{S}=H^{T}AH$. Replacing this into the objective function $J$ we can rewrite the optimization problem as
\begin{align}
& \argmin_{H\geq 0, H^{T}H=I}  tr (AA-2H^{T}AH H^{T}AH + H^{T}AH H^{T}AH ) \nonumber \\
  \equiv & \argmin_{H\geq 0, H^{T}H=I}  tr (AA-H^{T}AH H^{T}AH ) \nonumber \\
 \equiv & \argmax_{H\in \mathcal{H}_{+}^{N \times K}} \|H^{T}AH\|_F.
\label{OSNTF_obj}
\end{align}
We recognize the last term as $\sqrt{tr(APAP)}$, where $P=HH^{T}$ is the projection matrix onto the subspace defined by the columns of the matrix $H$.  Once we obtain $\hat{H}$ that optimizes the objection function in $J$, we can obtain $\hat{S}=\hat{H}^{T}A\hat{H}$ and
\begin{equation}
\hat{A}=\hat{H} \hat{S} \hat{H}^{T}=\hat{H}\hat{H}^{T} A \hat{H}\hat{H}^{T}=\hat{P}A\hat{P}.
\label{projection}
\end{equation}

We conclude this section by acknowledging that it is not possible to solve either of the optimization problems (\ref{SNMFoptim}) or (\ref{OSNTFoptim}) to obtain a global maximum, and the algorithms we use only have a convergence guarantee to a local optimum.

\subsection{Uniqueness}
While finding if an exact SNMF or OSNTF of order $K$ exists is NP-hard, it is worth investigating that given such a factorization exists, whether it is even possible to uniquely recover the parts or factors through non-negative matrix factorization. In other words is SNMF or OSNTF unique? In fact, a long standing concern about using NMF based procedures is their non-uniqueness in recovering the data generating factors under general settings. This issue has been investigated in detail in \citet{donoho03,laurberg08} and \citet{huang14}. The next lemma builds upon the arguments presented in \citet{laurberg08,huang14} and \citet{ding06} to show that SNMF is unique only up to an orthogonal matrix, and OSNTF is unique except for a permutation matrix when the rank of $A$ or $L$ is $K$.
\begin{lem}
For any $N \times N$ symmetric matrix $A$, if $rank(A)=K \leq N$, then the order $K$ \textit{exact} SNMF of $A$ is unique up to an orthogonal matrix, and the order $K$ \textit{exact} OSNTF of $A$ is unique up to a permutation matrix, provided the exact factorizations exist.
\label{lem:uniqueness}
\end{lem}
The proof of this lemma along with those of all other lemmas and theorems are given in the Appendix. We also have the following useful corollary for OSNTF which is also proved in the Appendix.

\begin{cor}
For any $N \times N$ symmetric matrix $A$, if $[H,S]$ is an \textit{exact} OSNTF of $A$, then each row of $H$ contains only one non-zero (positive) element.
\label{nonzero}
\end{cor}

\section{Motivation and connection with spectral clustering}

\subsection{Connections to invariant subspaces and projections}
We now connect SNMF and OSNTF to invariant subspaces of a linear transformation on a finite dimensional vector space. Suppose $[\hat{H},\hat{S}]$ is an OSNTF of order $K$ of the matrix $A$. Then by Equation (\ref{projection}), $\hat{A}=\hat{P}A\hat{P}$ is an at most rank $K$ matrix approximating $A$. By definition $\hat{A}$ has an exact OSNTF of order $K$.

Focusing on the \textit{exact} OSNTF factorization for the moment, we note that the factorization in (\ref{OSNTF}) of order $K \leq N$ can be equivalently written as
\begin{equation}
AH=HSH^{T}H=HS, \qquad H_{N \times K}\geq 0, S_{K \times K}\geq 0, H^{T}H =I. \label{invariant}
\end{equation}
Since $H$ has $K$ orthonormal columns, $rank(H)=K$. Consequently, if an OSNTF of order $K$ exists for $A$, then the columns of $H$ span a $K$-dimensional invariant subspace, $\mathcal{R}(H)$, of $A$. Moreover, since $H$ is orthogonal in this case, the columns of $H$ form an orthogonal basis for the subspace $\mathcal{R}(H)$. Every eigenvalue of $S$ is an eigenvalue of $A$ and the corresponding eigenvector is in  $\mathcal{R}(H)$. To see this note that if $x$ is an eigenvector of $S$ corresponding to the eigenvalue $\lambda$, then, $Sx=\lambda x$. Now, $AHx=HSx=\lambda Hx$ and hence $Hx$ is an eigenvector of $A$ and is in $\mathcal{R}(H)$. Moreover since in this case, $rank(A)=rank(S)$, $S$ contains all the non zero eigenvalues of $A$ as its eigenvalues.

Note that the projection matrix onto the column space of $H$, i.e., $\mathcal{R}(H)$, is given by $P=HH^{T}$. Since $AP=AHH^{T}=HSH^{T}=HH^{T}HSH^{T}=PA$, $\mathcal{R}(H)$ is also a reducing subspace of the column space of $A $ \citep{radjavi03,stewart}. Hence the following decomposition holds (called the spectral resolution of $A$) :
\begin{equation}
\begin{pmatrix}
H_1^{T} \\
H_2^{T}
\end{pmatrix} A (H_1\,  H_2) =
\begin{pmatrix}
S_{1} & 0 \\
0 & S_{2}
\end{pmatrix}
\label{specproj}
\end{equation}
where $H_1$ and $H_2$ are matrices whose columns span $\mathcal{R}(H)$ and its orthogonal complement respectively \citep{stewart}.

Reverting back to the approximate factorization, we notice that the optimization
problem in Equation (\ref{OSNTFoptim}) is to find the best projection of $A$ into an at most rank $K$ matrix $\hat{A}$ which has a non-negative invariant subspace. Note, here and subsequently, the ``best" approximation implies a matrix which minimizes the distance in Frobenius norm. The difference of this projection with the projection in spectral clustering through singular value decomposition \citep{mcsherry01,qr13} is that the projection in singular value decomposition ensures that the result is the best at most rank $K$ matrix approximating $A$, however it does not necessarily have a non-negative invariant subspace.
In that sense the OSNTF projection adds an additional constraint on the projection and consequently the resultant matrix is no longer the best at most rank  $K$ approximating matrix. In OSNTF, the non-negative invariant subspace $\mathcal{R}(H)$ is used for community detection. Hence in general, the discriminating subspace in OSNTF is different from the one used in spectral clustering.

We make a similar observation for SNMF. The order $K$ factorization defined in (\ref{SNMF}) can be written as
\begin{equation}
AH=HH^{T}H=HS' \qquad H\geq 0, S'\geq 0, \label{invariant2}
\end{equation}
where $S'=H^{T}H$ is clearly a positive semi-definite matrix. In addition, if we assume that the matrix $A$ is of rank $K$, then $rank(H) \geq rank(A)=K$. Hence the matrix $H$ is also of rank $K$ and has independent
columns. By the preceding argument, columns of $H$ span an invariant subspace of $A$ and the columns of $H$ is a basis (not orthogonal) for the subspace $\mathcal{R}(H)$. However $S'$ in this case is positive semi-definite and hence has only non-negative eigenvalues. Consequently the subspace spanned by the columns of $H$ only contains the subspaces associated with the non-negative eigenvalues. In contrast, since $S$ in OSNTF can have both positive and negative eigenvalues, that means $\mathcal{R}(H)$ contains subspaces associated with both positive and negative eigenvalues.

\subsection{Motivation through block diagonal matrix}

To motivate community detection with symmetric NMF methods, we start by looking into a special case where the graph is made of $K$ separate connected components, i.e., the adjacency matrix is block-diagonal. As a consequence the normalized Laplacian matrix $L$ is also block-diagonal. In this case we have very clear $K$ clusters in the graph. Note that the probability of connection within the blocks can vary arbitrarily and no special structure is assumed within the blocks. Spectral clustering was motivated in \citet{von07} and \citet{ng02} through a similar block-diagonal Laplacian matrix. The arguments in those papers were as follows. For the normalized Laplacian matrix of any undirected graph without multiple edges and self loops, the eigenvalues lie between $[-1,1]$. The blocks in the block diagonal matrix $L$ are also themselves normalized Laplacian matrices of the connected components of the graph. Since the spectrum of $L$ is a union of the spectra of $L_i$, the matrix $L$ has exactly $K$ eigenvalues of magnitude 1, each coming from one of the blocks. Hence selecting the eigenvectors corresponding to the largest $K$ eigenvalues of $L$ will select $K$ eigenvectors $\mathbf{x}_{L_i}$, each being the leading eigenvector of one of the blocks. Hence the subspace formed with those eigenvectors will naturally be discriminant for the cluster structure. A similar motivation for OSNTF is contained in the next lemma.
\begin{lem}
Let $L= \begin{pmatrix}
L_1 & &  \\
& \ddots & &   \\
  & & L_K
\end{pmatrix}_{N \times N}$ be a block-diagonal Laplacian matrix of a graph with $K$ connected components with sizes $m_{1}, \ldots,m_{K}$. There exists a one dimensional invariant subspace of block $L_i \in \mathcal{R}^{m_i \times m_i}$ with a non-negative basis $h_i \in \mathcal{R}^{m_i \times 1}$, for all $i$. Let $e_{L_i}$ denote an $N \times 1$ vector obtained by extending $h_i$ by adding $\sum_{t=1}^{i-1} m_{t}$ 0's at the top and $\sum_{t=i+1}^{K} m_{t}$ 0's at the bottom. Then the orthogonal non-negative basis $e_{L}=\{e_{L_1}, \ldots , e_{L_K} \}$ spans a $K$ dimensional invariant subspace of $L$.
\label{lem:invariant}
\end{lem}

The previous lemma shows that a block-diagonal $L$ has a $K$ dimensional invariant subspace with a non-negative basis matrix composed of vectors which are indicators for the blocks and consequently naturally discriminant. However it is not immediately clear if this subspace or the basis matrix is recovered by SNMF and OSNTF. The next theorem uses the Perron-Frobenius theorem to show that SNMF and OSNTF can indeed correctly recover the block memberships of the nodes from this Laplacian matrix. It turns out that in this ideal case of completely disconnected $K$ clusters, SNMF, OSNTF and spectral clustering use the same subspace for clustering.

\begin{thm}
Let $G_{N,K}$ be a graph with $N$ nodes and $K$ connected components. Both the SNMF and the OSNTF correctly recover the component memberships of the nodes from the block-diagonal normalized Laplacian matrix of this graph.
\label{thm:block}
\end{thm}

\section{Consistency of OSNTF for community detection}

We now turn our attention to more general adjacency and Laplacian matrices. The stochastic block model (SBM) is a well studied statistical model of a network with community structure. The $K$ block stochastic block model assigns to each node of a network, a $K$ dimensional community label vector which takes the value of $1$ in exactly one position and $0$'s everywhere else. Let $Z$ be a matrix whose $i$th row is the community label vector for the $i$th node. Given the community labels of the nodes, the edges between them are formed independently following a Bernoulli distribution with a probability that depends only on the community assignments, i.e., given community assignments there is stochastic equivalence among the nodes. A node is said to ``belong to" community $k$ if its vector of community labels has $1$ in the $k$th position. We further assume that there is at least one non-zero element in each column, i.e., each community has at least one node. The SBM can be written in the matrix form as
\begin{equation}
E(A)=\mathcal{A}=ZBZ^{T}, \quad B \in [0,1]^{K \times K}, \, Z \in \{0,1\}^{N \times K}, \label{SBM}
\end{equation}
where the matrix $B \geq 0$ is a $K\times K$ symmetric matrix of probabilities. We assume the matrix $B$ is of full rank, i.e., of rank $K$. We will refer to the matrix $\mathcal{A}$ as the population adjacency matrix. The population Laplacian matrix is defined from this adjacency matrix as $\mathcal{L}=\mathcal{D}^{-1/2}\mathcal{A} \mathcal{D}^{-1/2}$, where $\mathcal{D}$ is a diagonal matrix with the elements being $\mathcal{D}_{ii}=\sum_{j} \mathcal{A}_{ij}$. The matrix $\mathcal{L}$ under the $K$ class SBM defined above can be written as
\begin{equation}
\mathcal{L}=ZD_{B}^{-1/2}BD_{B}^{-1/2}Z^{T}=ZB_{L}Z^{T},
\label{SBML}
\end{equation}
 where $D_{B}=diag(BZ^{T}\mathbf{1}_{N}) \in R^{K \times K}$ with $\mathbf{1}_{N}$ being the vector of all ones in $\mathcal{R}^{N}$, is a diagonal matrix and $B_{L}=D_{B}^{-1/2}BD_{B}^{-1/2}$ \citep{rcy11}. The square root of $D_B$ and its inverse are well defined since the $K$ elements of the diagonal matrix $D_B$ are strictly positive.

Although the SBM is a well-studied model, it is not very flexible in terms of modeling real world networks. Many real world networks exhibit heterogeneity in the degrees of the nodes which the SBM fails to model. To remedy this, an extension of SBM for general degree distributions was proposed in \citet{kn11}, called the degree corrected stochastic block model (DCSBM). In our matrix terms the model can be written as
\begin{equation}
E(A)=\mathcal{A}=\Theta ZB'Z^{T} \Theta, \quad B' \in \mathcal{R}_{+}^{K \times K}, \, Z \in \{0,1\}^{N \times K}  ,\, \Theta \in \mathcal{R}_{+}^{N \times N}, \label{DCSBM}
\end{equation}
where $B'$ is a symmetric full rank matrix and $\Theta$ is an $N \times N$ diagonal matrix containing the degree parameters $\theta_{i}$ for the nodes as elements. Following \citet{kn11} we impose identifiability constraints $\sum_{\{i: Z_{iq}=1\}}\theta_{i}=1$ for each $q \in \{1,\ldots, K\}$. Note that $B'$ is not a matrix of connection probabilities any more. Instead the interpretation of $B'$ is that each entry $B^{'}_{ql}$ represents the expected number of edges between communities $q$ and $l$. The population Laplacian matrix for DCSBM can be obtained from the population adjacency matrix as,
\begin{equation}
\mathcal{L}=\Theta^{1/2} ZB^{'}_{L}Z^{T} \Theta^{1/2},  \label{DCSBML}
\end{equation}
where $B^{'}_L=D_{B}^{'-1/2}B^{'}D_{B}^{'-1/2}$, and $D^{'}_{B}$ is defined as $D^{'}_{B}=diag(\sum_{l}B^{'}_{1l}, \ldots, \sum_{l}B^{'}_{Kl})$. \citep{qr13}.

We prove that applying OSNTF to either the adjacency matrix $A$ or the Laplacian matrix $L$ is consistent for community detection in graphs generated from both the SBM and the DCSBM.  The analysis for consistency consists of several steps. We first assume that the population adjacency matrix (and the Laplacian matrix) is a $K$ class stochastic block model and demonstrate that it can be written as an OSNTF. Then we show that OSNTF can correctly recover the class assignments from this population adjacency (Laplacian) matrix. The observed sample adjacency (Laplacian) matrix is then viewed as a perturbed version of the population adjacency (Laplacian) matrix, and hence an approximate OSNTF algorithm through optimization can recover the class assignments with some errors. Finally we bound the proportion of errors by establishing uniform convergence of the observed objective function to the population objective function. The analysis for DCSBM is similar.

\subsection{Recovery}

\subsubsection{SBM}
The next lemma shows that the procedure OSNTF can recover the class assignments perfectly from the population adjacency matrix or the Laplacian matrix generated by the stochastic block model. Hence even though for any given matrix both proving the existence and evaluation of exact OSNTF is NP hard, if we know that the matrix is formed according to the stochastic block model, the methods can still recover true class assignments. We can then hope that the  methods can recover the true class assignments from the sample adjacency of Laplacian matrices with high probability as well.

A careful examination of Equation (\ref{SBM}) reveals that the model is ``almost" an OSNTF with the exception that the columns of $Z$ are orthogonal to each other, but not orthonormal. Hence $Z^{T}Z$ is a diagonal matrix, instead of a identity matrix. The next lemma exploits this relationship to establish that an OSNTF on $\mathcal{A}$ and $\mathcal{L}$ can correctly recover the class assignments.

\begin{lem}
There exists a diagonal ``scaling" matrix $Q =(Z^TZ) \in \mathcal{R}^{K \times K}$ with strictly positive entries such that $[\bar{H}=ZQ^{-1/2}, \bar{S}=Q^{1/2}BQ^{1/2}]$ and $[\bar{H}_L=ZQ^{-1/2}, \bar{S}_L=Q^{1/2}B_{L}Q^{1/2}]$ are the (unique up to a permutation matrix $P$) solutions to the OSNTF of $\mathcal{A}$ and $\mathcal{L}$ respectively under the $K$ block stochastic block model as defined in Equation (\ref{SBM}). Moreover,
$$Z_{i} Q^{-1/2} =Z_{j} Q^{-1/2} \iff Z_{i}=Z_{j}, $$ where $Z_{i}$ is the $i$th row of $Z$.
\label{lem:recovery1}
\end{lem}

The previous lemma shows that OSNTF of rank $K$ applied to the population adjacency or the Laplacian matrix of a SBM obtains factors $[\bar{H},\bar{S}]$ such that any two rows of $\bar{H}$ are equal if and only if the corresponding rows are equal in $Z$. Now assigning rows to communities on the basis of the largest entry in $\bar{H}$ as in Equation (\ref{assignment}) effectively means doing the same on rows of $Z$, which by definition will result into correct community assignments. However due to the ambiguity in terms of a permutation matrix $P$, the community labels can be identified only up to a permutation.

\subsubsection{DCSBM}

We now prove a parallel result on recovery of class assignments from the population adjacency and Laplacian matrices of DCSBM.

\begin{lem}
There exists diagonal ``scaling" matrices $Q=(Z^T\Theta^2Z), Q_L=(Z^T\Theta Z) \in \mathcal{R}^{K \times K}$ with strictly positive entries such that $[\bar{H}= \Theta ZQ^{-1/2}, \bar{S}=Q^{1/2}B'Q^{1/2}]$ and $[\bar{H}_L=\Theta^{1/2} ZQ_L^{-1/2}, \bar{S}_L=Q_L^{1/2}B^{'}_{L}Q_L^{1/2}]$ are the (unique up to a permutation matrix) solutions to the OSNTF of $\mathcal{A}$ and $\mathcal{L}$ respectively under the $K$ block DCSBM as defined in Equation (\ref{DCSBM}). Moreover, nodes $i$ and $j$ are assigned to the same community if and only if $Z_i =Z_j$.
\label{lem:recovery2}
\end{lem}

\subsection{Uniform convergence of objective function}
Although OSNTF can perfectly recover $Z$ from the population adjacency matrix $\mathcal{A}$ and the population Laplacian matrix $\mathcal{L}$, in practice we do not observe $\mathcal{A}$ or $\mathcal{L}$. Instead we observe the sample version (or perturbed version) of $\mathcal{A}$, the sample adjacency matrix $A$. The sample version of $\mathcal{L}$ can be obtained from $A$ by $L=D^{-1/2}AD^{-1/2}$.

To upper bound the difference between the observed perturbed version with the true population quantity for both the adjacency matrix and the Laplacian matrix, we reproduce Theorems 1 and 2 of \citet{chung11} in the following proposition.
\begin{prop}
 (\citet{chung11}) Let $A^{(N)} \in \{0,1\}^{N \times N}$ be a sequence of random adjacency matrices corresponding to a sequence of binary undirected  random graphs with $N$ nodes and population adjacency matrices $E(A^{(N)})=\mathcal{A}^{(N)} \in [0,1]^{N \times N}$. Let $L^{(N)}$ and $\mathcal{L}^{(N)}$ be the corresponding sample and population graph Laplacians respectively. Let $\Delta_N$ and $\delta_N$ denote the maximum and minimum expected degree of a node in the graph respectively. For any $\epsilon>0$, if $\Delta_N>\frac{4}{9} \log \left(\frac{2N}{\epsilon}\right)$ for sufficiently large $N$, then with probability at least $1-\epsilon$,
 $$\|A^{(N)}-\mathcal{A}^{(N)}\|_2 \leq  2\sqrt{ \Delta_N \log (2N/ \epsilon)}, $$
and for any $\epsilon'>0$, if $\delta_N > c(\epsilon') \log (N)$ for some constant $c(\epsilon')$, then with probability at least $1-\epsilon'$,
 $$\|L^{(N)}-\mathcal{L}^{(N)}\|_2 \leq  2\sqrt{\frac{3\log (4N/ \epsilon)}{\delta_N}}. $$
 \label{Abound}
\end{prop}

The observed sample adjacency matrix $A$ may not have an exact OSNTF. In that case, let the optimization problem in (\ref{OSNTFoptim}) or equivalently in (\ref{OSNTF_obj}), obtain a solution $[\hat{H},\hat{S}]$ as OSNTF of $A$. The matrix approximating $A$ is then $\hat{A}=\hat{H}\hat{S}\hat{H}^{T}$ and we assign the nodes to the communities using the matrix $\hat{H}$.

We denote the objective function in the optimization problem of (\ref{OSNTF_obj}) as $ F(A,H)=\|H^{T}AH\|_F $. This is a function of the adjacency matrix $A$ and the degree corrected community assignment matrix $H$. We can define a corresponding ``population" version of this objective function with the population adjacency matrix as $F(\mathcal{A},H)=\|H^{T}\mathcal{A}H\|_F $. The corresponding observed and population versions for the Laplacian matrix are defined by $ F(L,H_L)=\|H_L^{T}LH_L\|_F $ and $F(\mathcal{L},H_L)=\|H_L^{T}\mathcal{L}H_L\|_F $ respectively. In the next lemma we prove two uniform convergences: we show that for any $H \in \mathcal{H}_{+}^{N \times K} $, $F(A,H)$ converges to $F(\mathcal{A},H)$ and for any $H_L \in \mathcal{H}_{+}^{N \times K} $, $F(L,H_L)$ converges to $F(\mathcal{L},H_L)$ with proper scaling factors.

To determine the scaling factors, we look at the the growth rate of the maximized population versions $F(\mathcal{A},\bar{H})$ and $F(\mathcal{L},\bar{H}_L)$ under SBM and DCSBM. We assume that for both SBM and DCSBM, the probability of connections between the nodes grows as $\mathcal{A}_{ij} \asymp p$, uniformly for all $i,j$. We define $d=\sum_{i,j}\mathcal{A}_{ij}/N$ as the expected average degree of the nodes. Clearly, $d \asymp d_{i} \asymp Np$.
Then under both the SBM and the DCSBM we have,
\begin{align}
   F(\mathcal{A},\bar{H})  =\|\bar{H}^T\mathcal{A}\bar{H}\|_F & =\|\bar{S}\|_F \nonumber \\
   & =  \sqrt{tr( \bar{S}\bar{H}^T\bar{H}\bar{S}\bar{H}^T\bar{H})} \quad \text{(since } \bar{H}^T\bar{H}=I \text{)}
   \nonumber \\
   & = \sqrt{tr (\mathcal{A}\mathcal{A})} =\|\mathcal{A}\|_F = \sqrt{\sum_{i,j} \mathcal{A}_{ij}^2}
\asymp \sqrt{N^2 p^2} \asymp Np \asymp d.
\label{scaling}
\end{align}


Based on a similar argument, for $F(\mathcal{L},\bar{H}_L)$ we have under both the SBM and the DCSBM,
\begin{equation}
    F(\mathcal{L},\bar{H}_L)= \|\bar{S}_{L}\|_F
     = \|\mathcal{L}\|_F
    =\sqrt{\sum_{i,j} \frac{\mathcal{A}_{ij}^2}{\mathcal{D}_{ii} \mathcal{D}_{jj}}}
     \asymp \sqrt{N^2 \left(\frac{p}{Np}\right)^2}  \asymp 1.
     \label{scalingL}
\end{equation}


With this guidance, we use $\|\mathcal{A}\|_F$ as a scaling factor for the convergence of $F(A,H)$ and no scaling factor for the convergence of $F(L,H_L)$.
\begin{lem}
Consider the settings of Proposition \ref{Abound} and assume the minimum expected degree of the network $\delta$ grows as $\omega((\log N)^2)$. For any $H \in \mathcal{H}_{+}^{N \times K} $ we have,
\begin{equation}
|F(L,H)-F(\mathcal{L},H)| \overset{p}{\to} 0,
\end{equation}
provided $K=o(\delta/\log N)$, while,
\begin{equation}
\frac{1}{\|\mathcal{A}\|_F}|F(A,H)-F(\mathcal{A},H)| \overset{p}{\to} 0,
\end{equation}
provided $K=o(\|\mathcal{A}\|_F/\log N)$ and the maximum expected degree of the network $\Delta \asymp \|\mathcal{A}\|_F$.
\label{lem:convergence}
\end{lem}

While the above lemma does not assume any growth rate on $\|\mathcal{A}\|_F$, in order to interpret it we assume uniform growth on the probability of connections, and use the rate obtained in Equation (\ref{scaling}), i.e., $\|\mathcal{A}\|_F \asymp d$. In the dense case, where the expected degree of the nodes grows linearly with the number of nodes, we have $d \asymp N$, and in the sparse case, where we only assume the expected degree grows faster than $(\log N)^2$, we have $d = \omega ((\log N)^2)$. Then for OSNTF on the adjacency matrix of a dense expected network, where the expected degree of the nodes grows linearly with the number of nodes, we have $\|\mathcal{A}\|_F \asymp N$ and we can let the number of communities grow as $K=o(N/\log N)$. In the sparse case of poly-logarithmic growth on expected degree, where we only assume the expected degree grows faster than $(\log N)^2$, we have $\|\mathcal{A}\|_F = \omega ((\log N)^2)$ and we can let $K$ grow as $K=o(\log N)$.

We have a similar observation for OSNTF on Laplacian matrix as well. In the dense case when the minimum expected degree $\delta$ grows as $O(N)$, the condition on the growth of $K$ again turns out to be $K=o(N/\log N)$ and in the sparse case where $\delta=\omega((\log N)^2)$, we have $K=o(\log N)$.


\subsection{Characterizing mis-clustering}
 Although OSNTF can perfectly recover $Z$ from $\mathcal{A}$, in practice we obtain the orthogonal matrix $\hat{H}$ from the observed adjacency matrix $A$ instead of obtaining $\bar{H}$. Consequently, community assignment using the largest entry in each row of $\hat{H}$ as in Equation (\ref{assignment}) will lead to some error. We quantify the error through a measure called mis-clustering rate which, given a ground truth community assignment and a candidate community assignment, computes the proportion of nodes for which the assignments do not agree. However, due to the ambiguity in terms of permutation of community labels, we need to minimize the proportion over the set of all possible permutation of labels. Let $\bar{e}$ denote the ground truth and $\hat{e}$ denote a candidate assignment. Then we define the mis-clustering rate
 \[
 r = \frac{1}{N} \inf_{\Pi} d_H(\bar{e},\Pi(\hat{e})),
 \]
where  $\Pi(\cdot)$ is a permutation of the labels and $d_H(\cdot,\cdot)$ is the Hamming distance between two vectors.

\subsubsection{SBM}
The next result relates the error with the difference of the matrices $\hat{H}$ and $\bar{H}$.

\begin{lem}
Let $Z$ be the true community assignment matrix for a network generated from the stochastic block model and $Q=Z^{T}Z$. Let $(\hat{H},\hat{S})$ be the factorization of the adjacency matrix as in (\ref{OSNTFoptim}). Then any mis-clustered node $i$ must satisfy
\begin{equation}
\|\hat{H}_{i}-\bar{H}_{i}P\| > \frac{1}{\sqrt{N_{max}}},
\label{MSBM}
\end{equation}
where $\hat{H}_{i}$ and $\bar{H}_{i}$ denote the $i$th row of the matrices $\hat{H}$ and $\bar{H}$ respectively, $P$ is a permutation matrix, and  $N_{max}=\max_{k \in \{1, \ldots, K\}} Q_{kk}$, i.e., the population of the largest block. This is also the necessary condition for mis-clustering node $i$ in OSNTF of the Laplacian matrix.
\label{lem:MSBM}
\end{lem}

The next theorem is our main result for OSNTF under SBM, which uses this characterization of misclustering along with the bounds obtained in previous lemmas to bound the misclustering rate.

\begin{thm}
Let $G$ be a graph generated from a $K$ class SBM with parameters $(Z,B)$ as in Equation (\ref{SBM}). Define $\lambda^{\mathcal{A}}$ and $\lambda^{\mathcal{L}}$ as the smallest non-zero (in absolute value) eigenvalues of $\mathcal{A}$ and $\mathcal{L}$ respectively. Let $A, L \in R^{N \times N}$ be the adjacency and Laplacian matrices of the graph respectively, and define $r_A$ and $r_L$ as the mis-clustering rate for community detection through OSNTF of $A$ and $L$ respectively. If the conclusion of Lemma \ref{lem:convergence} on uniform convergence of $F(A,H)$ and $F(L,H)$ to $F(\mathcal{A},H)$ and $F(\mathcal{L},H)$ respectively holds, and the following conditions on $N_{\max}$, $\lambda^{\mathcal{A}}$ and $\lambda^{\mathcal{L}}$ hold:
(a) $N_{\max} \asymp \frac{N}{K}$,
(b) $\lambda^{\mathcal{A}} \geq 4 \|\mathcal{A}\|_F^{1/2} \left( \Delta \log (2N/\epsilon)\right/K)^{1/4}$,  and (c) $\lambda^{\mathcal{L}} \geq 4 \|\mathcal{L}\|_F^{1/2}  \left(\frac{3 \log (4N/\epsilon)}{K\delta}\right)^{1/4}$,
then
\begin{equation}
r_A \overset{p}{\to} 0,  \quad \quad \text{and} \quad \quad  r_L \overset{p}{\to}  0.
\end{equation}
\label{th:SBM}
\end{thm}

Note that under the assumption of uniform growth of connection probabilities, from Equation (\ref{scalingL}), we have $\|\mathcal{L}\|_F \asymp 1$. Then condition (c) in Theorem \ref{th:SBM} reduces to $\lambda^{\mathcal{L}} \gtrsim 4\left(\frac{3 \log (4N/\epsilon)}{K\delta}\right)^{1/4}$, which is similar to the condition on the same quantity in Theorem 4.2 of \citet{qr13}, and is closely related to the signal to noise ratio of the SBM.

\subsubsection{DCSBM}
We first prove a lemma connecting the event of mis-clustering with the difference between matrices $\hat{H}$ and $\bar{H}$, and matrices $\hat{H}_L$ and $\bar{H}_L$ for $A$ and $L$ respectively .
\begin{lem}
For a network generated from the DCSBM with parameter $(\Theta,Z,B)$ as in Equation (\ref{DCSBM}), let $(\hat{H},\hat{S})$ be the factorization of the adjacency matrix as in (\ref{OSNTF}). Then a necessary condition for any node $i$ to be mis-clustered is
\begin{equation}
\|\hat{H}_{i}-\bar{H}_{i}P\| \geq m,
\label{MDCSBM}
\end{equation}
where $m=\min_{i \in \{1, \ldots, N\}} \theta_{i}/\sqrt{(Z^{T}\Theta^{2} Z)_{kk}}$ with $k$ being the community to which the node $i$ truly belongs. The corresponding necessary condition for the OSNTF in Laplacian matrix is
\begin{equation}
\|\hat{H}_{L,i}-\bar{H}_{L,i}P\| \geq m',
\end{equation}
with $m'=\min_{i \in \{1, \ldots, N\}} \sqrt{\theta_{i}/(Z^{T}\Theta Z)_{kk}}$.
\label{lem:MDCSBM}
\end{lem}

The next theorem is our main result for OSNTF under DCSBM, which bounds the mis-clustering rate.

\begin{thm}
Let $G$ be a graph generated from a $K$ class DCSBM with parameters $(\Theta,Z,B)$ as in (\ref{DCSBM}). Define $\lambda^{\mathcal{A}}$ and $\lambda^{\mathcal{L}}$ as the smallest (in absolute value) eigenvalues of $\mathcal{A}$ and $\mathcal{L}$ respectively. Let $A, L \in R^{N \times N}$ be the adjacency and Laplacian matrices of the graph respectively, and define $r_A$ and $r_L$ as the mis-clustering rate for community detection through OSNTF of $A$ and $L$ respectively. If the conclusion of Lemma \ref{lem:convergence} on uniform convergence of $F(A,H)$ and $F(L,H)$ to $F(\mathcal{A},H)$ and $F(\mathcal{L},H)$ respectively holds, and the following conditions on $m$, $m'$, $\lambda^{\mathcal{A}}$ and $\lambda^{\mathcal{L}}$ hold:
(a) $m, m' \asymp \sqrt{\frac{K}{N}}$,
(b) $\lambda^{\mathcal{A}} \geq 4 \|\mathcal{A}\|_F^{1/2} \left(\Delta \log (2N/\epsilon)/K \right)^{1/4}$,  and (c) $\lambda^{\mathcal{L}} \geq 4 \|\mathcal{L}\|_F^{1/2} \left(\frac{3 \log (4N/\epsilon)}{K\delta}\right)^{1/4}$,
then
\begin{equation}
r_A \overset{p}{\to} 0,  \quad \quad \text{and} \quad \quad  r_L \overset{p}{\to}  0.
\end{equation}
\end{thm}

The proof is similar to that of the SBM case and will be omitted. To check that condition (a) is reasonable, in addition to the assumption of uniform growth on the probability of connections, we assume the population of the communities $N_q$ grows as $N_q \asymp \frac{N}{K}$ for all $q \in \{1,\ldots, K\}$. Then under the DCSBM we have $\theta_{i} \theta_{j} B'_{ql} =\mathcal{A}_{ij} \asymp p$ for all $i,j$ such that $Z_{iq}=1$ and $Z_{jl}=1$. Since by the identifiability constraint, $\sum_{i : Z_{iq}=1}\theta_{i} =1$, we have $\theta_{i} \asymp \frac{K}{N}$, $(Z^T \Theta Z)_{qq} \asymp 1$ and $(Z^T \Theta^2 Z)_{qq} \asymp \frac{K}{N}$.

\subsection{Application to four parameter SBM}
We apply Theorem \ref{th:SBM} to the four parameter SBM, which is a special case of SBM, parameterized by four parameters, $a,b,s,K$ \citep{rcy11,qr13}. The probability of connection within a block is $a$ for all blocks and the probability of connection between nodes from different blocks is $b$ for all block pairs. The connection probabilities are assumed to remain constant as $N$ grows. The number of nodes within each block is $s$ (hence all blocks are of the same size) and $K$ is the number of blocks. Then we have $N_{\max}=s=N/K$, $\delta \asymp N$, $\|\mathcal{L}\|_F \asymp 1$ and $\lambda^{\mathcal{L}}=\frac{1}{K(b/(a-b))+1} \asymp 1/K$ \citep{rcy11}. Then condition (c) of Theorem \ref{th:SBM} holds if $K=o((N/\log N)^{1/3})$. Hence from Theorem \ref{th:SBM}, the misclustering rate $r_{L} \rightarrow 0$, and we have consistent community detection. We also note that for the four parameter SBM case, condition (c) of Theorem \ref{th:SBM} and condition (a) of Theorem 4.2 in \citet{qr13} lead to the same constraint, namely, $K=o((N/\log N)^{1/3})$.

\section{Simulation Results}
In this section we generate networks from both the SBM and the DCSBM and evaluate the performance of NMF based approaches along with a few spectral methods applied to the normalized Laplacian matrix of the networks. The spectral methods we consider are the spectral clustering (Spectral) \citep{rcy11,lei14}, the regularized spectral clustering (Reg. Spectral) \citep{qr13} and regularized spectral clustering without projection (Spectral-wp)\citep{qr13}. We conduct three experiments generating data from the SBM for the first two and from the DCSBM for the last one. The clustering quality of a partition is evaluated by measuring its agreement with a known ground truth using the normalized mutual information (NMI) criterion. The NMI is an information theoretic measure of agreement between two partitions that takes value between 0 and 1, with higher values indicating better agreement between the partitions. All results are averaged over 32 simulations.

\begin{figure}[hp]
\centering{}
\begin{subfigure}{0.5\textwidth}
\centering{}
\includegraphics[width=1\linewidth]{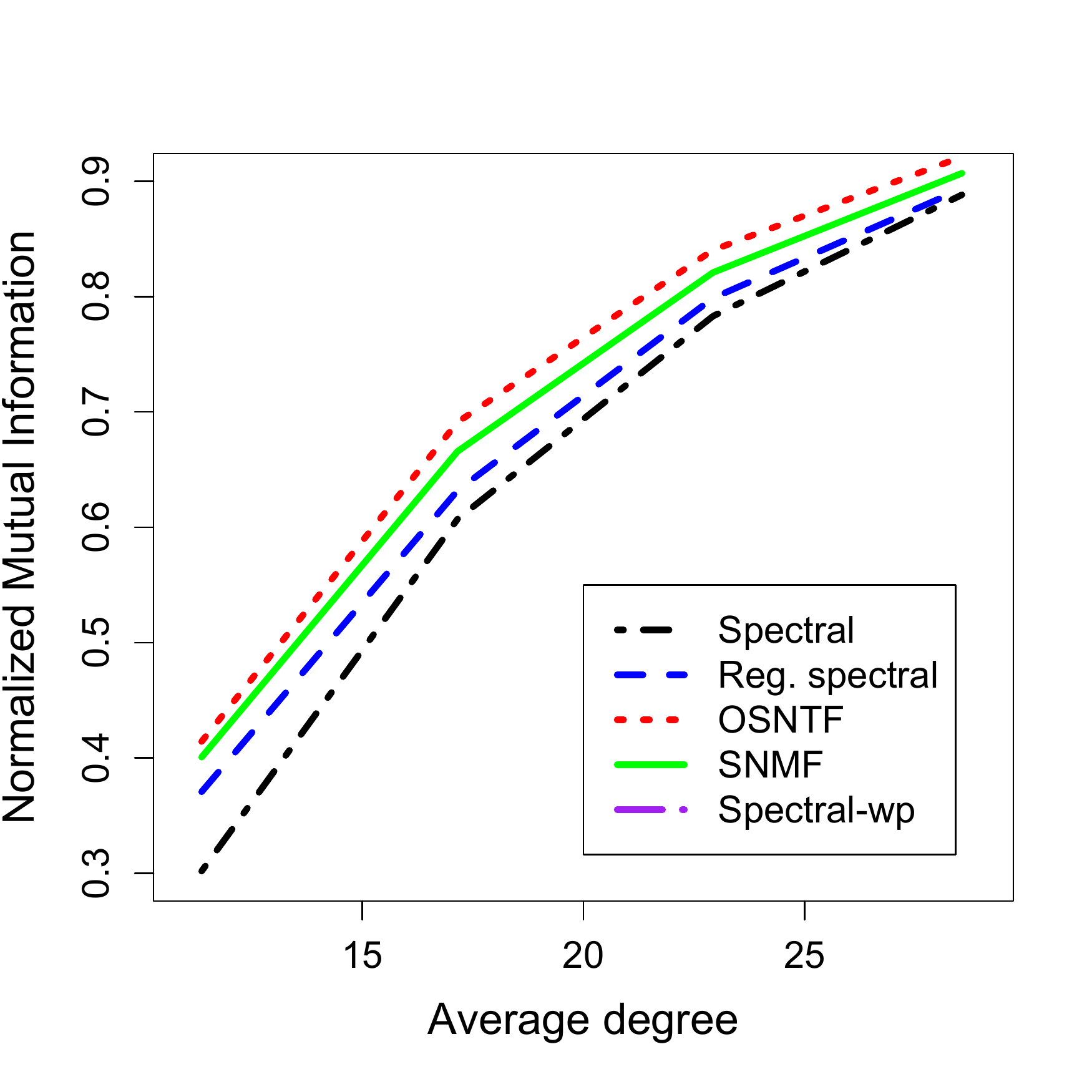}
\end{subfigure}%
\begin{subfigure}{0.5\textwidth}
\centering{}
\includegraphics[width=1\linewidth]{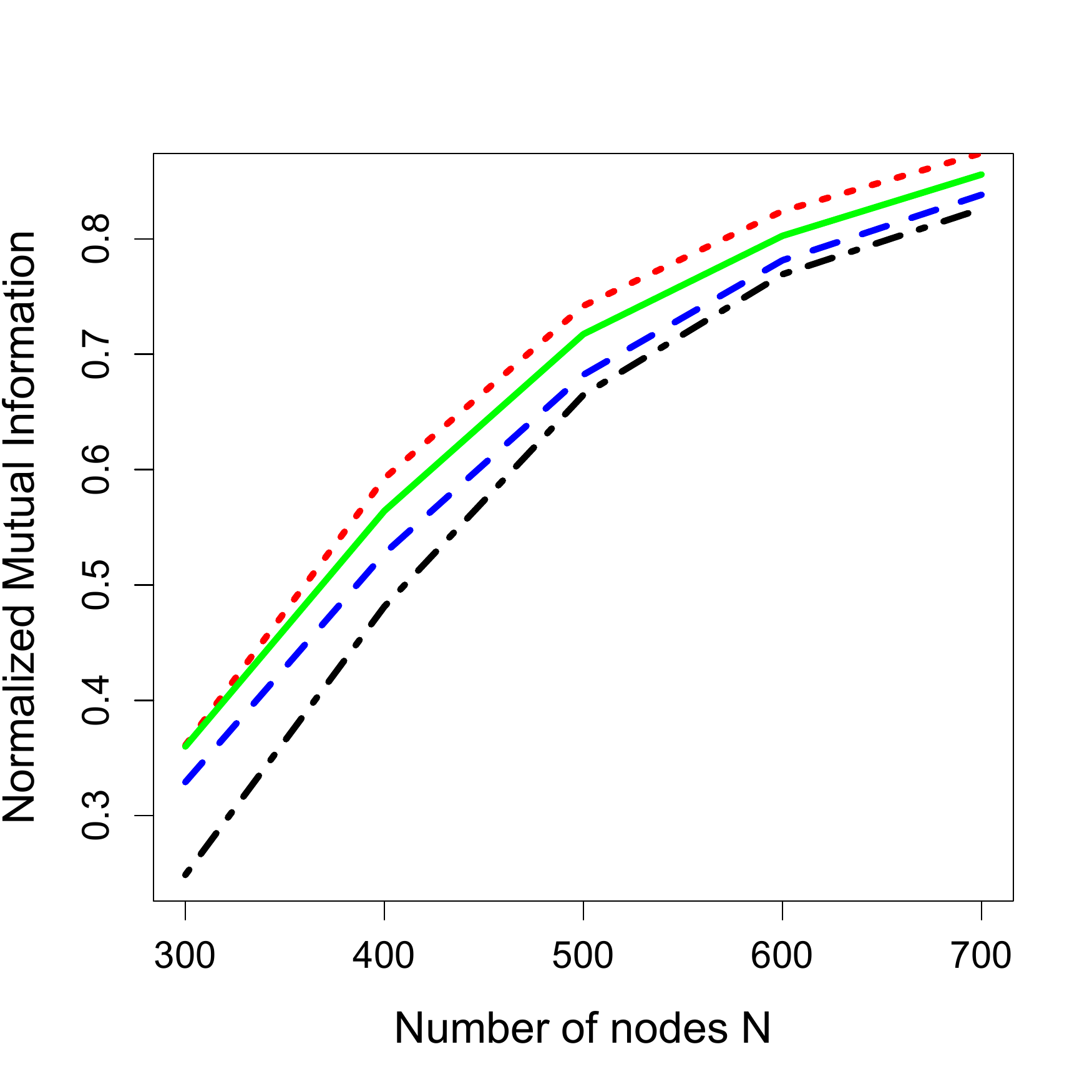}
\end{subfigure}
\begin{center} (a) \hspace{150pt} (b) \end{center}
\begin{subfigure}{0.5\textwidth}
\centering{}
\includegraphics[width=1\linewidth]{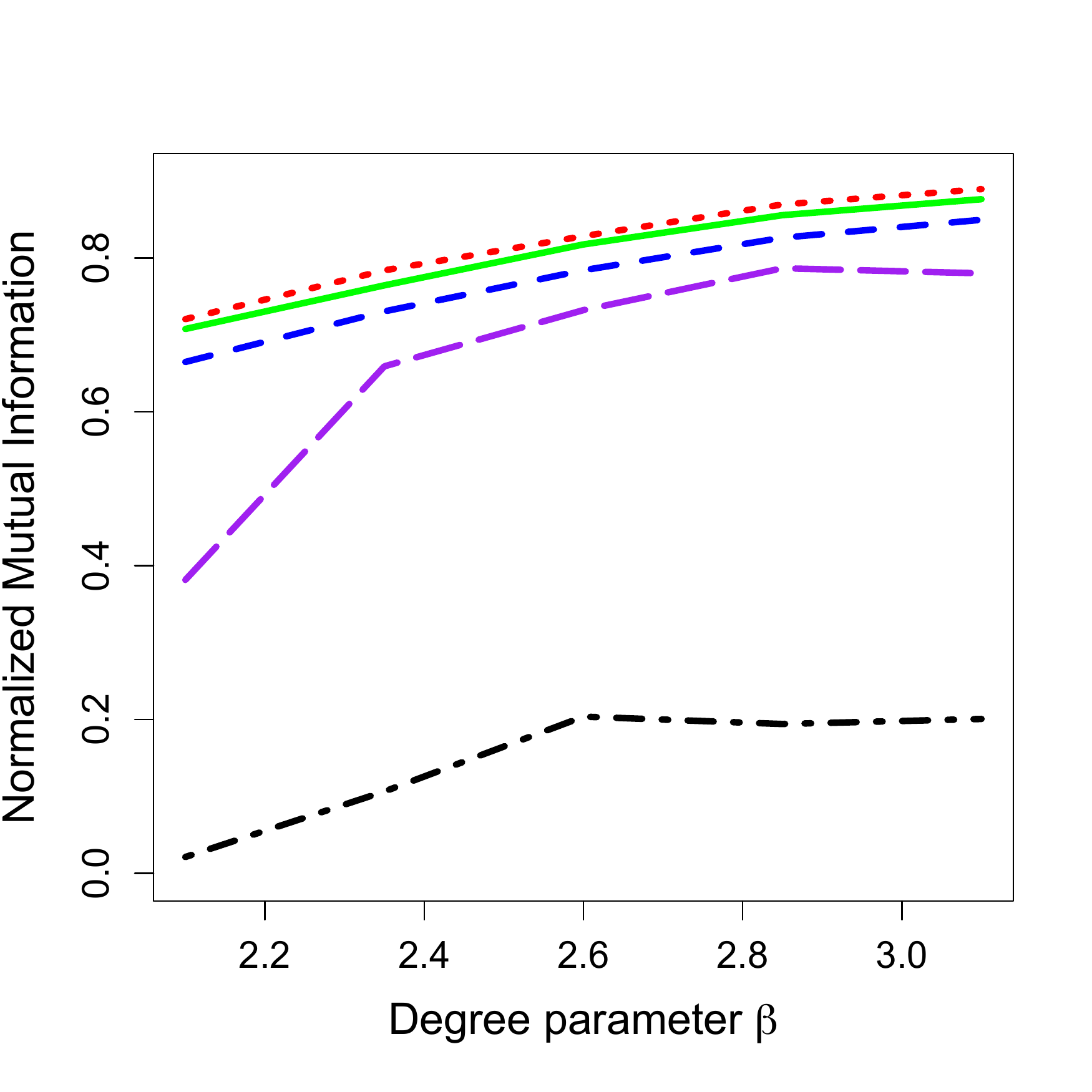}
\end{subfigure}
\begin{center} (c) \end{center}
\caption{\small Comparison of the performance of various methods for three simulation settings: (a) SBM with $N=800$, $K=3$ and increasing average degree, (b) SBM with $K=3$ and increasing number of nodes, and (c) DCSBM with $N=600$, $K=3$ and decreasing degree heterogeneity. The legend in Figure (a) is common to all figures.}
\label{simulation_figure}
\end{figure}

\subsection{SBM : increasing degrees}
We generate data from the SBM with 3 clusters and 800 nodes. The signal to noise ratio, defined as the ratio of the diagonal to off-diagonal elements, is kept fixed at around 3, while we increase the average degree of the network from 10 to 30. This simulation is designed to test the robustness of the methods for sparse graphs where node degrees are relatively low. The results are presented in Figure \ref{simulation_figure}(a). We notice that SNMF, OSNTF and regularized spectral clustering perform similarly with the NMF based methods having slightly higher NMI compared to the regularized spectral clustering throughout. The usual spectral clustering without any regularization performs slightly worse in low degree graphs (i.e. sparse graphs). Clearly this drawback of spectral clustering is not shared by SNMF and OSNTF as they perform well without any regularization. This also relieves us from the problem of choosing a suitable regularization parameter.

\subsection{SBM : increasing nodes}

For this experiment we generate data from SBM with 3 clusters and fixed connection probability matrix but vary the number of nodes (and as a consequence the average degree of network also gets varied). The aim of this study is to determine the number of nodes required by different methods to attain a comparable NMI. We again fix the signal to noise ratio at 3. The results presented in Figure \ref{simulation_figure}(b) look quite similar to the previous case. We notice that both SNMF and OSNTF consistently perform better than spectral clustering and regularized spectral clustering. Also the spectral clustering performs poorly when the number of nodes is low and regularization helps in that case.

\subsection{DCSBM : varying degree parameter}

In our last experiment we generate data from a DCSBM with 3 cluster and 600 nodes. The degree parameter is generated from a power law distribution with lower bound parameter $x_{min}=1$ and shape parameter $\beta$. We increase the shape parameter from $2.1$ to $3.1$ in steps of $0.25$. A smaller $\beta$ leads to greater degree heterogeneity and hence increasing the parameter gradually makes the DCSBM more similar to a SBM. We again keep the signal to noise ratio at 3 and the average degree of the networks generated is around 30. The results are presented in Figure \ref{simulation_figure}(c). Here we see that the (unregularized) spectral clustering completely breaks down in the presence of degree heterogeneity and recovers slowly as the parameter $\beta$ increases. Spectral clustering without projection (but with regularization) performs poorly when $\beta$ is 2.1 but recovers significantly as $\beta$ increases. This observation is consistent with that of \citet{qr13}. SNMF, OSNTF and regularized spectral clustering are robust against degree heterogeneity with the NMF based methods once again consistently outperforming regularized spectral clustering. This simulation study indicates that both SNMF and OSNTF perform well under the DCSBM without any modification.

A comparison of all the methods in terms of the number of times a method performs the best over all simulations is reported in Table \ref{tab:winner}. Clearly in all the three simulation scenarios, the NMF based methods turn out to perform overwhelmingly better compared to spectral and regularized spectral clustering. Moreover, OSNTF performs the best in almost $80 \%$ of the total simulations under consideration.

\begin{table}[h]
\protect\caption{Comparison of the methods in terms of the number of times a method performs the best in the simulations}

\centering
\begin{tabular}{ccccccc}
\hline
Simulation & SNMF & OSNTF & Spectral & Reg. Spectral & Spectral-wp & Total  \tabularnewline
\hline
SBM : incr. deg. & 22 & \bf{134}  & 1 & 3& $-$ & 160 \tabularnewline
SBM : incr. node & 10 & \bf{117}  & 1 & 0 & $-$ & 128 \tabularnewline
DCSBM  & 28 & \bf{128}  & 0 & 2 & 2 & 160 \tabularnewline
\hline
\end{tabular}

\label{tab:winner}
\end{table}

\section{Real data analysis}
In this section we apply SNMF and OSNTF to a few popular real network datasets with known ground truth and compare their performance with competing methods. All methods are applied to the normalized Laplacian matrix of the networks.

\subsection{Political blogs data}
We analyze the political blogs dataset collected by \citet{adamic05}. The dataset comprises of 1490 political blogs during US presidential election with the directed edges indicating hyperlinks. We consider the largest connected component of the graph comprising of 1222 nodes and convert it into an undirected graph by assigning an edge between two nodes if there is an edge between the two in any direction. The resultant network has an average degree of 27. This dataset with the above mentioned preprocessing was also analyzed by \citet{kn11,amini13,qr13,joseph13,j15,zlz12,gao15}, etc. for community detection, and is generally considered as a bench mark for evaluating algorithms. The ground truth community assignments partitions this network into two groups, liberal and conservative, according to the political affiliations or leanings of the blogs. Table \ref{tab:polblogs} summarizes the performance of the proposed NMF based methods along with that of spectral clustering, regularized spectral clustering  \citep{qr13} and SCORE method \citep{j15} in terms of the number of nodes mis-clustered. All methods except the regular spectral clustering perform similarly and correctly clusters approximately $95 \%$ of the nodes. Moreover, both the NMF based methods outperform the state of the art methods \citep{gao15}, e.g., regularized spectral clustering and SCORE. Note that our results for regularized spectral clustering (using the average node degree 27 as regularization parameter) is different from \citet{qr13}, where it was reported to miscluster $80 \pm 2$ nodes out of 1222. We believe this is primarily because \citet{qr13} use a different construction to convert the directed network into undirected network and obtain one with average node degree as 15. Our construction matches with that of \citet{amini13,joseph13,j15} and \citet{gao15}.

\begin{table}[h]
\protect\caption{Comparison of NMF with other methods in terms of the number of nodes mis-classified and the NMI with the ground truth in the political blogs dataset}

\centering
\begin{tabular}{ccccccc}
\hline
Measure & SNMF & OSNTF & Spectral & Reg. Spectral & SCORE \tabularnewline
\hline
 Misclustered & \bf{54} & \bf{54}  & 551 & 64 & 58* \tabularnewline
NMI & \bf{0.7455} & \bf{0.7455}  & 0.0523 & 0.7133 & 0.725* \tabularnewline
\hline
\end{tabular}

* results taken from \citet{j15}.
\label{tab:polblogs}
\end{table}

\subsection{Karate club dataset}

The second dataset we analyze is another well studied benchmark network, the Zachary' karate club data \citep{zachary77}. The data consist of friendship patterns among the 34 members of a karate club in a US university. Shortly after the data were collected the group split into two subgroups. Those sub-groups are our ground truth. This network has also been extensively studied in the literature \citep{ng04,bc09,j15}. Both SNMF and OSNTF clusters all the nodes in two communities correctly. We also note that both spectral and  regularized spectral clustering cluster the nodes perfectly while SCORE mis-clusters one node \citep{j15}.

\subsection{Dolphins dataset}

We consider an undirected social network of associations among 62 dolphins living in Doubtful Sound, New Zealand, curated by \citet{dolphin2}. Similar to the previous data set, during the course of the study it was observed that a well connected dolphin coded as SN100 left the group and this resulted into a split of the group into two subgroups. These subgroups consisting of the remaining 61 dolphins constitute our ground truth communities and we apply various community detection methods on this dataset. Both SNMF and OSNTF mis-cluster one node (SN89). In comparison the spectral clustering mis-clusters 11 nodes and the regularized spectral clustering mis-clusters 2 nodes.

\section{Discussions}
In this paper we have used a factorization of the Laplacian matrix with non-negativity and orthogonality constraints for community detection in complex networks. The proposed method was shown to be asymptotically consistent for community detection in graphs generated from the stochastic block model and the degree corrected stochastic block model. This method is quite similar to spectral clustering and attempts to estimate the same discriminating subspace as spectral clustering for a block-diagonal Laplacian matrix that corresponds to a graph with $K$ connected components. However, for more general graphs the two methods obtain very different invariant subspaces for discrimination. Our simulations show that this method outperforms the spectral clustering in a wide variety of situations. In particular, for sparse graphs and for graphs with high degree heterogeneity, this method does not suffer from some of the issues spectral clustering faces. While it is clear from Eckart-Young theorem that spectral clustering uses the best $K$ dimensional subspace that represents the data, the subspace may not be the best discriminating subspace for clustering. How does the subspace obtained by OSNTF compare with that obtained by spectral clustering as a discriminating subspace for community detection under different types of graphs is an important question that needs to be explored further.

While we have focused here primarily on OSNTF, a future course of research would be to study SNMF for community detection in graphs generated from SBM and DCSBM. As mentioned in the introduction, SNMF has been previously applied for graph clustering in \citet{wang11,kuang12} and \citet{yang12}. A major difficulty in proving consistency is however that the \textit{exact} SNMF appears to fail to recover the true community assignments from the population version of the adjacency or the Laplacian matrix unless the matrix $B$ in the definition of SBM is a completely positive matrix, i.e., it can be written as $B=KK^T$, where $K$ is a non-negative matrix. However such an assumption will be difficult to verify as determining if a matrix is completely positive is NP hard.

\section{Appendix}

\subsection{Proof of Lemma \ref{lem:uniqueness} and Corollary \ref{nonzero}}

\begin{proof}
We first prove Corollary \ref{nonzero}. If there are two non-zero elements in a row of $H$, say $H_{ik},H_{il} >0$, then their product would be a positive quantity. However since the columns of $H$ are orthonormal, $\sum_{i}H_{ik}H_{il} =0$. This would require the product $H_{i'k}H_{i'l}$ to be negative for some other $i'$. However, this is not possible since all the elements of $H$ are non-negative as well.

Now we prove Lemma \ref{lem:uniqueness}.
Suppose the SNMF as defined in Equation (\ref{SNMF}) is not unique and there is another factorization of $A$ as $A=H'H'^{T}$. If the matrix $A$ is of rank $K$, then since $rank(H) \geq rank(A)=K$ and $rank(H) \leq \min(N,K)$, we have $rank(H)=rank(H')=K$. The subspace spanned by the columns of matrices $H$ and $H'$ are the same, i.e., $span(H)=span(A)=span(H')$. Hence there exists an orthogonal change of basis matrix $Q \in \mathcal{R}^{K \times K}$ such that $H'=HQ^{T}$ and $Q^{T}Q=I$. Moreover this is the only possible source of non-uniqueness in $H$ \citep{laurberg08}. Consequently if $H$ is a solution, then all other SNMF solutions are of the form $HQ^{T}$ for any orthogonal matrix $Q$. Note that unlike asymmetric NMF, where the ambiguity is in terms of an arbitrary change of basis matrix $Q$ and its inverse, for SNMF the matrix $Q$ can only be an orthogonal matrix \citep{huang14}.

Applying the above arguments to OSNTF of order $K$ we have if $H$ and $S$ are a solution, then so is $H'=HQ$ and $S'=R^{T}SR$ where $QR^{T}=I$. Moreover, if $rank(A)$ is $K$, then both $H$ and $H'$ span the same subspace and must be related through an orthogonal change of basis matrix. Consequently, this is the only source of non-uniqueness.
However for OSNTF even this ambiguity of an orthogonal matrix is not possible due to the orthogonality and non-negativity constraints except for permutation matrices. If $HQ$ is a solution, then $HQ$ must have orthonormal columns, i.e., $(HQ)^{T}HQ=I$ which implies $Q^{T}Q=I$. However, except $Q=I$ or a permutation matrix, at least one element of $Q$ must be negative in order for it to be an orthogonal matrix \citep{ding06}. However, if an element of $Q$, say $Q_{kl}$, is negative, then $(HQ)_{il}=\sum_{k} H_{ik}Q_{kl}<0$ for all rows $i$ of $H$ such that the only non-zero element in the row is in the $k$th place (Note that such a row always exists, since no column of the rank $K$ matrix $H$ can be all 0's). This will make $HQ$ contain at least one negative element, which violates the non-negativity constraint. Hence the factorization is unique up to permutations.

\end{proof}

\subsection{Proof of Lemma \ref{lem:invariant}}

\begin{proof}
Let $A$ denote the block-diagonal adjacency matrix of the graph with $K$ connected components. Since each of the blocks in $A$, denoted by $A_{i}$, is an adjacency matrix of a connected component of the graph, they are irreducible non-negative matrices, and the same is true for the Laplacian matrix $L$ (Theorem 2.2.7 of \citep{berman79}). Hence by Perron-Frobenius Theorem, for each of these connected components, there exists one positive real eigenvalue (called the Perron root, $\rho (L_{i})$) and the corresponding eigenvector has all positive entries. Moreover, the Perron root is simple, unique and the largest eigenvalue of $L_i$ for each of the blocks (Theorem 1.2 of \citep{chang08}). Hence the eigenspace spanned by the eigenvector is one-dimensional. Let $h_i \in \mathcal{R}^{m_i \times 1}$ denote this eigenvector for block $L_{i}$. Then $h_i$ is a non-negative basis for a one-dimensional invariant subspace of block $L_i \in \mathcal{R}^{m_i \times m_i}$. Now each of $\rho (L_{i})$ is also an eigenvalue of $L$ since the spectrum of $L$ is the union of the spectra of $L_i$. Since $e_{L_i}$ denotes an $N \times 1$ vector obtained by extending $h_i$ by adding $0$'s in the place of the remaining blocks as described in the statement of the lemma, it is also an invariant subspace of $L$ corresponding to the eigenvalue $\rho (L_{i})$. Hence the orthogonal non-negative basis $\{e_{L_1}, \ldots , e_{L_k} \}$ spans a $K$ dimensional invariant subspace of $L$.

\end{proof}

\subsection{Proof of Theorem \ref{thm:block}}
\begin{proof}
Let $L=U \Sigma U^{T}$ be the eigen-decomposition of $L$ where $\Sigma$ is a diagonal matrix containing the eigenvalues in the diagonal in decreasing order and  $U \in \mathcal{R}^{N\times N}$ is an orthogonal matrix of the corresponding eigenvectors. Spectral clustering then proceeds by stacking the eigenvectors corresponding to the top $K$ eigenvalues of $L$ into a matrix $U_K \in \mathcal{R}^{N \times K}$. As discussed in Section 3.2, in this case each of the largest $K$ eigenvalues is the largest eigenvalue of one of the blocks and the columns of $U_K$ are $\mathbf{x}_{L_i}$ augmented with 0's in the place of the remaining blocks. We note from the proof of Lemma \ref{lem:invariant} that for each block, the eigenvalue is then Perron root of that block ($\rho(L_{i})$), and the matrix $U_K$ is made of $e_{L_i}$, and consequently, $U_K \geq 0$.

Now by Eckart-Young Theorem, $ \min_{rank(L') \leq K} \|L-L'\|_F$ is minimized by $\hat{L}=U_{K}\Sigma_{K}U_{K}^{T}$, where the diagonal matrix $\Sigma_K$ contains the top $K$ eigenvalues of $L$ in decreasing order in its diagonal \citep{eckart36}. In other words, $U_{K}\Sigma_{K}U_{K}^{T}$ is the best at most $rank(K)$ approximation to $L$. Note that both SNMF and OSNTF of order $K$ will approximate $L$ by a matrix with rank $K$ or less.

Since $U_{K} \geq 0$ and $U_{K}^{T}U_{K}=I$, it is clear that the factorization  $\hat{L}=U_{K}\Sigma_{K}U_{K}^{T}$ is an OSNTF. This is also a SNMF since $U_{K}\Sigma_{K}U_{K}^{T}=U_{K}\Sigma_{K}^{1/2} \Sigma_{K}^{1/2} U_{K}^{T}=H_KH_{K}^{T}$ with $H_{K}=U_{K}\Sigma_{K}^{1/2} \geq 0$. Note that $\Sigma_{K}^{1/2}$ exists since all the diagonal elements of $\Sigma_{K}$ are 1's. Hence, $\hat{L}$ is the approximating matrix of rank $K$ in the solution of the optimization problem for both SNMF and OSNTF. Consequently, $[U_K,\Sigma_K]$ and $U_{K}\Sigma_{K}^{1/2}$ are the OSNTF and SNMF of order $K$ respectively for $L$.

By construction of $U_K$, for each row $i$ of $U_K$, $k=\argmax_{j \in \{1,\ldots, K\}} (U_K)_{ij}$ will indicate the block to which node $i$ belongs to. Hence both SNMF and OSNTF will cluster the nodes perfectly.

\end{proof}

\subsection{Proof of Lemma \ref{lem:recovery1}}
\begin{proof}
We have by definition of the stochastic block model,
$$ \mathcal{A}=ZBZ^{T}, \quad Z^{T}Z=Q^{K\times K}, \quad det(B)\neq 0, $$
where $Q$ is a diagonal matrix whose diagonal elements $\{Q_{11}, \ldots, Q_{KK}\}$ are the population of the different blocks. Clearly an OSNTF of order $K$ applied to the matrix $\mathcal{A}$ will not yield the matrices $Z$ and $B$, since $Z^{T}Z \neq I$. However, notice that
\begin{equation}
\mathcal{A}=ZBZ^{T}=Z(Z^{T}Z)^{-1/2}(Z^{T}Z)^{1/2}B(Z^{T}Z)^{1/2}(Z^{T}Z)^{-1/2}Z^{T}=\bar{H}\bar{S}\bar{H}^{T},
 \label{sbmdecompA}
\end{equation}
where $\bar{H}=Z(Z^{T}Z)^{-1/2}=ZQ^{-1/2}$ and $\bar{S}=(Z^{T}Z)^{1/2}B(Z^{T}Z)^{1/2}=Q^{1/2}BQ^{1/2}$. Since we assume all the communities in the stochastic block model have at least one member, all the elements of the diagonal matrix $Q$ are strictly positive quantities. Hence both the square root matrix $Q^{1/2}$ and its inverse exist and are well defined. Clearly, $ \bar{H}^{T}\bar{H}=I$ and $ \bar{H},\bar{S} \geq 0$. Hence, $[\bar{H},\bar{S}]$ is an OSNTF of rank $K$ for $\mathcal{A}$. Any other OSNTF of rank $K$ for the matrix $\mathcal{A}$ is unique up to a permutation matrix $P$ by Lemma \ref{lem:uniqueness}.

For the result on $\mathcal{L}$, we have from Equation (\ref{SBML}),
\begin{equation}
\mathcal{L}=ZB_{L}Z^{T}=ZQ^{-1/2} Q^{1/2}B_{L}Q^{1/2}Q^{-1/2} Z^{T}.
\label{sbmdecompL}
\end{equation}
Hence, following the preceding argument, an OSNTF of rank $K$ applied to the matrix $\mathcal{L}$ will recover the factor matrices as $\bar{H}_L=ZQ^{-1/2}$ and $\bar{S}_L=Q^{1/2}B_LQ^{1/2}$ unique up to a permutation matrix $P$.

Since $Q^{1/2}$ and $Q^{-1/2}$ exist, $Z_{i} Q^{-1/2}=Z_{j} Q^{-1/2} \iff Z_{i}=Z_{j} $ in both cases.

\end{proof}

\subsection{Proof of Lemma \ref{lem:recovery2}}

\begin{proof}
The population adjacency matrix of the DCSBM, as in Equation (\ref{DCSBM}), is
\begin{align}
\mathcal{A} &= \Theta ZBZ^{T} \Theta \nonumber \\
&=\Theta Z (Z^{T} \Theta^{2} Z)^{-1/2}(Z^{T}\Theta^{2}Z)^{1/2}B(Z^{T}\Theta^{2}Z)^{1/2}(Z^{T}\Theta^{2}Z)^{-1/2}Z^{T} \Theta \nonumber \\
&=\bar{H}\bar{S}\bar{H}^{T},
 \label{dcsbmdecompA}
\end{align}
where $\bar{H}=\Theta Z(Z^{T}\Theta^{2}Z)^{-1/2}=\Theta ZQ^{-1/2}$ and $\bar{S}=(Z^{T}\Theta^{2}Z)^{1/2}B(Z^{T}\Theta^{2}Z)^{1/2}=Q^{1/2}BQ^{1/2}$. Note that the matrix $Q=(Z^{T}\Theta^{2}Z)=(\Theta Z)^{T}(\Theta Z) \in R^{K \times K}$, is a diagonal matrix. Clearly all the elements are strictly positive and hence the matrix admits both a square root and an inverse. We compute
 $$ \bar{H}^{T}\bar{H}=(Z^{T}\Theta^{2}Z)^{-1/2} (Z^{T} \Theta^{2} Z) (Z^{T}\Theta^{2}Z)^{-1/2} =I,$$
 and $ \bar{H},\bar{S} \geq 0$. Hence, $[\bar{H},\bar{S}]$ is an OSNTF of rank $K$ for $\mathcal{A}$ under DCSBM. Any other OSNTF of rank $K$ for the matrix $\mathcal{A}$ is unique up to a permutation matrix $P$ by Lemma \ref{lem:uniqueness}.

Since both $Q^{1/2}$ and $Q^{-1/2}$ exist, we have $Z_{i}Q^{-1/2}=Z_{j}Q^{-1/2}$ if and only if $Z_{i}=Z_{j}$. Moreover, since $Z_i$ contains only one non-zero element, say at position $k$, and $Q$ is a diagonal matrix, $(ZQ^{-1/2})_{i}$ also has only one non-zero element, whose position within the row is also $k$. Now,
$\arg \max_{j} \bar{H}_{ij} =\arg \max_{j} \theta_{i}(ZQ^{-1/2})_{ij} = \arg \max_{j} (ZQ^{-1/2})_{ij}$. Hence, nodes $i$ and $j$ will be assigned to the same community if and only if $Z_i =Z_j$.

Similarly for $\mathcal{L}$, from Equation (\ref{DCSBML}),
\begin{align}
\mathcal{L} &=\Theta^{1/2} ZB_{L}Z^{T} \Theta^{1/2} \nonumber\\
&=\Theta^{1/2} Z (Z^{T} \Theta Z)^{-1/2}(Z^{T}\Theta Z)^{1/2}B_{L}(Z^{T}\Theta Z)^{1/2}(Z^{T}\Theta Z)^{-1/2}Z^{T} \Theta^{1/2} \nonumber \\
&=\bar{H}_L\bar{S}_L\bar{H}_{L}^{T},
\label{dcsbmdecompL}
\end{align}
where $\bar{H}_L=\Theta^{1/2} Z(Z^{T}\Theta Z)^{-1/2}=\Theta^{1/2} ZQ_{L}^{-1/2}$ and $\bar{S}_L=(Z^{T}\Theta Z)^{1/2}B_L(Z^{T}\Theta Z)^{1/2}=Q_{L}^{1/2}B_{L}Q_{L}^{1/2}$. We note that the matrix $Q_{L}=Z^{T}\Theta Z \in R^{K \times K}$ is also a diagonal matrix with strictly positive diagonal entries and hence both square root and inverse are well defined. Since $\bar{H}_{L}^{T}\bar{H}_{L}=I$ and $ \bar{H}_{L},\bar{S}_{L} \geq 0$, $[
\bar{H}_{L},\bar{S}_{L}]$ is an OSNTF of rank $K$ for the matrix $\mathcal{L}$. As before, this is unique up to a permutation matrix $P$.

The proof for the second part is identical to the previous case with $\mathcal{A}$.
\end{proof}

\subsection{Proof of Lemma \ref{lem:convergence}}

\begin{proof}
We have for any $\epsilon>0$ and $\Delta =\omega(\log N)$,
\begin{align*}
\frac{1}{\|\mathcal{A}\|_F}|F(A,H)-F(\mathcal{A},H)| &= \frac{1}{\|\mathcal{A}\|_F} | \, \|H^{T}AH\|_F - \|H^{T}\mathcal{A}H\|_F | \\
& \leq \frac{1}{\|\mathcal{A}\|_F} \|H^{T}AH - H^{T}\mathcal{A}H\|_F  \\
& \leq \frac{1}{\|\mathcal{A}\|_F} \sqrt{K} \|H^{T}AH - H^{T}\mathcal{A}H\|_2 \\
& = \frac{1}{\|\mathcal{A}\|_F} \sqrt{K} \|H^{T}(A - \mathcal{A})H\|_2 \\
& \leq \frac{1}{\|\mathcal{A}\|_F} \sqrt{K} \|H\|^2_2 \|A-\mathcal{A}\|_2 \\
& \leq \frac{2 \sqrt{K \Delta \log (2N/ \epsilon)}}{\|\mathcal{A}\|_F},
\end{align*}
with probability $1-\epsilon$. The second line follows from the triangle inequality property of the Frobenius norm. The third line is due to the fact that $(H^{T}AH - H^{T}\mathcal{A}H)$ is a $K \times K$ matrix and the equivalence of norm relation, $\|X\|_F \leq \sqrt{\text{rank} (X)} \|X\|_2 $. The fifth line is due to the property of spectral norm that $\|ABC\|_2 \leq \|A\|_2\|B\|_2\|C\|_2$, while the sixth line follows from Proposition \ref{Abound} and the fact that $\|H\|^2_2=\lambda_{\max}(H^TH)=\lambda_{\max}(I_{K})=1$.

Hence under the assumptions that $\Delta \geq \delta =\omega (\log N)^2$, $\Delta \asymp \|\mathcal{A}\|_F$ and $K=o(\|\mathcal{A}\|_F/\log N)$ we have,
\[
\frac{1}{\|\mathcal{A}\|_F}|F(A,H)-F(\mathcal{A},H)|\overset{p}{\to} 0.
\]

Similarly for the objective function on the Laplacian matrix, we have for any $H \in \mathcal{H}_{+}^{N \times K} $ with $\delta =\omega (\log N)^2$,
\[
|F(L,H)-F(\mathcal{L},H)| \leq 2 \sqrt{\frac{3 K \log (4N/ \epsilon)}{\delta}}
\]
with probability $1-\epsilon$ for any $\epsilon>0$. The right hand side once again converges to 0 provided $K=o(\delta/\log N)$.
\end{proof}

\subsection{Proof of Lemma \ref{lem:MSBM}}

\begin{proof}
Since $\hat{H}$ is an exact OSNTF of $\hat{A}$, by Corollary \ref{nonzero}, each row of $\hat{H}$ has one non-zero element. If $\bar{H}_{ik}=(ZQ^{-1/2}P)_{ik} > 0$, then a correct assignment for row $i$ would require $\hat{H}_{ik} > 0$. This implies if node $i$ is incorrectly assigned, then
\begin{align*}
\|\hat{H}_{i}-\bar{H}_{i}P\|^2 & =\|\hat{H}_{i}-Z_{i}Q^{-1/2}P\|^2 =\|\hat{H}_{i}\|^2 + \|Z_{i}Q^{-1/2}P\|^2 \\
& \geq \|Z_{i}Q^{-1/2}P\|^2=\frac{1}{Q_{kk}} \geq \frac{1}{N_{max}}.
\end{align*}
Hence, every mis-clustered node $i$ must have  $\|\hat{H}_{i}-Z_{i}Q^{-1/2}P\|$ at least as large as $\frac{1}{\sqrt{N_{max}}}$, and a difference of less than $\frac{1}{\sqrt{N_{max}}}$ is a sufficient condition for correct clustering. The matrix $\bar{H}_L= ZQ^{-1/2}$ is the same for OSNTF in Laplacian matrix as it is for OSNTF in adjacency matrix, and hence the necessary condition for mis-clustering is also $\|\hat{H}_{L,i}-\bar{H}_{L,i}P\| \geq \frac{1}{N_{max}}$ .
\end{proof}

\subsection{Proof of Theorem \ref{th:SBM}}

\begin{proof}
First note that by equivalence in Equation (\ref{OSNTF_obj}), the $H$ which  maximizes $F(\mathcal{A},H)$ also minimizes $J=\|\mathcal{A}-HSH^{T}\|_F$, with $S=H^{T}\mathcal{A}H$. From Lemma \ref{lem:recovery1}, it immediately follows that $\bar{H}$ maximizes  $F(\mathcal{A},H)$ and $\bar{H}_L$ maximizes $F(\mathcal{L},H_L)$ for the SBM up to the ambiguity of permutation matrix $P$. Using Lemma \ref{lem:MSBM}, if the misclustering rate $r_A \geq \eta$ for some $\eta>0$, then
\[
\|\hat{H}-\bar{H}P\|_F^2 = \sum_{i} \|\hat{H}_i-\bar{H}_iP\|^2 \geq \sum_{i:\ i \text{ is misclustered}}  \|\hat{H}_i-\bar{H}_iP\|^2 \geq \frac{N \eta}{N_{\max}}.
\]
In other words the event $\{ r_A \geq \eta \} \subset \{ \|\hat{H}-\bar{H}P\|_F \geq \sqrt{\frac{N \eta}{N_{\max}}} \}$. Since $F(\mathcal{A},H)$ is uniquely maximized by $\bar{H}P$ for some permutation matrix $P$, we have $\frac{1}{\|\mathcal{A}\|_F}F(\mathcal{A},\bar{H}P) \geq \frac{1}{\|\mathcal{A}\|_F}F(\mathcal{A},\hat{H}) + 2 \tau$ whenever $\|\hat{H}-\bar{H}P\|_F \geq \sqrt{\frac{N \eta}{N_{\max}}}$ for some $\tau>0$. The result on mis-clustering rate follows provided $\tau$ is large enough, in particular $\tau = \frac{2 \sqrt{K \Delta \log (2N/ \epsilon)}}{\|\mathcal{A}\|_F}$,
\begin{align*}
P(r_A \geq \eta) & \leq P\left(\|\hat{H}-\bar{H}P\|_F \geq \sqrt{\frac{N \eta}{N_{\max}}} \right) \\
& \leq P \left[ \frac{1}{\|\mathcal{A}\|_F}F(\mathcal{A},\bar{H}P) \geq \frac{1}{\|\mathcal{A}\|_F}F(\mathcal{A},\hat{H}) + 2 \tau \right] \\
& = P \Bigg [ \left\{\frac{1}{\|\mathcal{A}\|_F}F(\mathcal{A},\bar{H}P) \geq \frac{1}{\|\mathcal{A}\|_F}F(\mathcal{A},\hat{H}) + 2 \tau \right \} \\
& \quad \quad \cap \left \{ \frac{1}{\|\mathcal{A}\|_F}F(A,\hat{H}) \geq \frac{1}{\|\mathcal{A}\|_F}F(A,\bar{H}P) \right \} \Bigg ] \\
 & \leq P \Bigg [ \left \{\frac{1}{\|\mathcal{A}\|_F}|F(\mathcal{A},\bar{H}P) - F(A,\bar{H}P)| \geq \tau \right\} \\
 & \quad \quad \cup \left \{ \frac{1}{\|\mathcal{A}\|_F}|F(\mathcal{A},\hat{H}) - F(A,\hat{H})| \geq \tau \right\} \Bigg ] \\
 & \leq P \Bigg [\frac{1}{\|\mathcal{A}\|_F}|F(\mathcal{A},\bar{H}P) - F(A,\bar{H}P)| \geq \tau \Bigg] \\
 & \quad \quad + P\Bigg[ \frac{1}{\|\mathcal{A}\|_F}|F(\mathcal{A},\hat{H}) - F(A,\hat{H})| \geq \tau \Bigg ] \\
 & \rightarrow 0.
\end{align*}
The third line follows from the fact that $F(A,\hat{H}) \geq F(A,\bar{H}P)$ since $\hat{H}$ is the maximizer of $F(A,H)$ and the last line follows from the proof of Lemma \ref{lem:convergence}.

The following lemma uses the celebrated Davis-Kahan Perturbation Theorem \citep{dk70} to show that $\tau$ satisfies the condition required for application of Lemma \ref{lem:convergence}.

\begin{lem}
Under the notations of Theorem \ref{th:SBM}, we have
\[
 F(\mathcal{A},\bar{H})-F(\mathcal{A},\hat{H})  \geq \frac{(\lambda^{\mathcal{A}})^2 \|\hat{H}-\bar{H}P\|_F^2}{4\|\mathcal{A}\|_F},
\]
and
\[
F(\mathcal{L},\bar{H}_L)-F(\mathcal{L},\hat{H}_L)  \geq \frac{(\lambda^{\mathcal{L}})^2 \|\hat{H}_L-\bar{H}_LP\|_F^2}{4\|\mathcal{L}\|_F}.
\]
\label{lem:dk}
\end{lem}

By assumption (a) $N_{\max} \asymp N/K$, we have $\frac{N}{N_{\max}} \asymp K$. Consequently, $\|\hat{H}-\bar{H}P\|_F^2 \gtrsim K$. Then Lemma \ref{lem:dk} along with assumption (b), i.e., $\lambda^{\mathcal{A}} \geq 4 \|\mathcal{A}\|_F^{1/2} ( \Delta \log (2N/\epsilon)/K)^{1/4}$ ensure that $\tau$ can be chosen as $\frac{2 \sqrt{K \Delta \log (2N/ \epsilon)}}{\|\mathcal{A}\|_F}$.

The result for $r_L$ follows by repeating the same arguments. Since, $F(\mathcal{L},H_L)$ is uniquely maximized by $\bar{H}_LP$, we have $F(\mathcal{L},\bar{H}_LP) \geq F(\mathcal{L},\hat{H}_L) + 2 \tau_L$, whenever $\|\hat{H}_L-\bar{H}_LP\|_F \geq \sqrt{\frac{N \eta}{N_{\max}}}$ for some $\eta>0$ and $\tau_L>0$. Lemma \ref{lem:dk} along with assumption (c) $\lambda^{\mathcal{L}} \geq 4 \|\mathcal{L}\|_F^{1/2} \left(\frac{3 \log (4N/\epsilon)}{K \delta}\right)^{1/4}$ ensure that $\tau_L$ can be chosen as $2 \sqrt{\frac{3 K \log (4N/ \epsilon)}{\delta}}$ and the uniform convergence result from Lemma \ref{lem:convergence} can be applied to obtain $P(r_L \geq \eta) \rightarrow 0$ for any $\eta>0$.
\end{proof}

\subsection{Proof of Lemma \ref{lem:MDCSBM}}

\begin{proof}
Following the previous arguments for the case of SBM in Lemma \ref{lem:MSBM}, if node $i$ is incorrectly assigned, then
\begin{align*}
\|\hat{H}_{i}-\bar{H}_{i}P\|^2 & =\|\hat{H}_{i}-\theta_{i} Z_{i} Q^{-1/2}P\|^2 =\|\hat{H}_{i}\|^2 + \|\theta_{i} Z_{i} Q^{-1/2}P\|^2 \\
& \geq \|\theta_{i} Z_{i} Q^{-1/2}P\|^2=\frac{\theta^2_{i}}{(Z^T\Theta^{2} Z)_{kk}} \geq m^2.
\end{align*}
For OSNTF of the Laplacian matrix, this necessary condition for mis-clustering becomes
$$ \|\hat{H}_{L,i}-\bar{H}_{L,i}P\|^2 = \|\hat{H}_{i}-\theta^{1/2}_{i} Z_{i} Q_L^{-1/2}P\|^2 \geq \frac{\theta_{i}}{(Z^T \Theta Z)_{kk}} \geq (m')^2.$$
\end{proof}

\subsection{Proof of Lemma \ref{lem:dk}}
\begin{proof}
Let $S_1= \hat{H}^{T}\mathcal{A}\hat{H}$ and $\mathcal{A}_1=\hat{H} S_1 \hat{H}^{T}$. Then $F(\mathcal{A},\bar{H})=\|\bar{H}^{T}\mathcal{A} \bar{H}\|_F= \|\bar{S}\|_F$ and $F(\mathcal{A},\hat{H})=\|\hat{H}^{T}\mathcal{A} \hat{H}\|_F= \|S_1\|_F$. Moreover, $[\hat{H},S_1]$ is an exact OSNTF of the matrix $\mathcal{A}_1$.

From the discussion in Section 3.1, the columns of $\bar{H}$ and $\hat{H}$ span reducing subspaces of $\mathcal{A}$ and $\mathcal{A}_1$ respectively. We can then look at the matrix $\mathcal{A}$ as a perturbed version of the matrix $\mathcal{A}_1$ and use the Davis-Kahan Perturbation Theorem \citep{dk70} to relate the difference between the subspaces $\mathcal{R}(\hat{H})$ and $\mathcal{R}(\bar{H})$ with the difference between $\mathcal{A}_1$ and $\mathcal{A}$. In the next proposition we first reproduce the perturbation theorem mentioned in Theorem 3.4, Chapter 5 of \citet{stewart} in terms of canonical angles between subspaces. Note that for any matrix $A$, $\Lambda(A)$ denotes the set of its eigenvalues. For two subspaces $\mathcal{E}$ and $\mathcal{F}$, the matrix $\Theta(\mathcal{E},\mathcal{F})$ is a diagonal matrix that contains the canonical angles between the subspaces in the diagonal. See \citet{stewart}, and \citet{vu13} for more details on canonical angles. We use $\sin\Theta(\mathcal{E},\mathcal{F})$ to denote the matrix that applies sine on every element of $\Theta(\mathcal{E},\mathcal{F})$.

\begin{prop}
(\citet{stewart}) Let the columns of $H_{1}^{N \times K}$ span a reducing subspace of the matrix $\mathcal{B}$, and let the spectral resolution of $\mathcal{B}$ as defined by Equation (\ref{specproj}) be
\begin{equation}
\begin{pmatrix}
H_1^{T} \\
H_2^{T}
\end{pmatrix} \mathcal{B} (H_1,  H_2) =
\begin{pmatrix}
K_{1} & 0 \\
0 & K_{2}
\end{pmatrix},
\end{equation}
where $(H_1, H_2)$ is an orthogonal matrix with $H_{1} \in \mathcal{R}^{N \times K}$, and $K_{1} \in \mathcal{R}^{K \times K} $ and $K_{2} \in \mathcal{R}^{(N-K) \times (N-K)}$ are real symmetric matrices. Let $X \in \mathcal{R}^{N \times K}$ be the analogous quantity of $H_1$ in the perturbed matrix $B$, i.e., $X$ has orthonormal columns and there exists a real symmetric matrix $M \in \mathcal{R}^{K \times K}$ such that $BX=XM$. Define $E= \mathcal{B}-B$. Then $R=\mathcal{B} X-XM=EX$. If  $\delta = \min_{\lambda_1 \in \Lambda(K_2), \lambda_2 \in \Lambda(M)} |\lambda_1 -\lambda_2 | >0 $, then
$$\|\sin \Theta ( \mathcal{R}(H_1), \mathcal{R}(X))\|_F \leq \frac{\|R\|_F}{\delta} \leq \frac{\|\mathcal{B} -B\|_F}{\delta}. $$
\label{daviskahan}
\end{prop}

To use the proposition in our context, let $\mathcal{B}=\mathcal{A}_1$, $B=\mathcal{A}$, $H_1=\hat{H}$, $X=\bar{H}$. Then we have $K_1=S_1$ and $M= \bar{S}$. Since $S_1$ contains all the non-zero eigenvalues of $\mathcal{A}_1$ (Section 3.1), in this case $\Lambda(K_2)$ contains only 0's. On the other hand $\Lambda(M)$ contains all the non-zero eigenvalues of $\mathcal{A}$. Consequently, $\delta = \min_{\lambda_1 \in \Lambda(K_2), \lambda_2 \in \Lambda(M)} |\lambda_1 -\lambda_2 | =\lambda^{\mathcal{A}}$.

By Proposition 2.2 of \citet{vu13} there exists a $K$ dimensional orthogonal matrix $O$ such that
\begin{equation}
\frac{1}{2} \|\hat{H}-\bar{H}O\|^2_F \leq \| \sin \Theta(\mathcal{R}(\hat{H}),\mathcal{R}(\bar{H}))\|^2_F \leq \frac{\|\mathcal{A}-\mathcal{A}_1\|_F^2}{(\lambda^{\mathcal{A}})^2}.
\label{basisrel}
\end{equation}

Next note that,
\begin{align*}
    \|\mathcal{A}-\mathcal{A}_1\|_F^2 & = \|\mathcal{A}\|_F^2 + \|\mathcal{A}_1\|_F^2 -2 tr(\mathcal{A}\mathcal{A}_1) \\
    & = \|\bar{S}\|_F^2 + \|S_1\|_F^2 - 2 tr(\mathcal{A}\hat{H}\hat{H}^T\mathcal{A}\hat{H}\hat{H}^T)\\
    & = \|\bar{S}\|_F^2 + \|S_1\|_F^2- 2 tr(\hat{H}^T\mathcal{A}\hat{H}\hat{H}^T\mathcal{A}\hat{H}) \\
    & = \|\bar{S}\|_F^2 + \|S_1\|_F^2- 2 tr(S_1S_1) \\
    & = \|\bar{S}\|_F^2 -  \|S_1\|_F^2.
\end{align*}
Hence from Equation (\ref{basisrel}) we have,
\[
\frac{1}{2}\|\hat{H}-\bar{H}P\|_F^2 \leq  \frac{\|\bar{S}\|_F^2 -  \|S_1\|_F^2}{(\lambda^{\mathcal{A}})^2} = \frac{(\|\bar{S}\|_F -  \|S_1\|_F)(\|\bar{S}\|_F + \|S_1\|_F)}{(\lambda^{\mathcal{A}})^2}.
\]
This implies
\[
\|\bar{S}\|_F -  \|S_1\|_F \geq \frac{(\lambda^{\mathcal{A}})^2 \|\hat{H}-\bar{H}P\|_F^2}{2(\|\bar{S}\|_F + \|S_1\|_F)}.
\]
Now from Equation (\ref{scaling}) we have $\|\bar{S}\|_F = \|\mathcal{A} \|_F $, and $\|\bar{S}\|_F =F(\mathcal{A},\bar{H}) \geq F(\mathcal{A},\hat{H})=\|S_1\|_F$. Hence $\|\bar{S}\|_F$ dominates the sum in the denominator. Replacing the denominator by $4\|\mathcal{A}\|_F$ we have the desired bound. The proof is identical for the result on Laplacian matrix.

\end{proof}

\bibliography{nmf}
\end{document}